\documentclass[12pt]{elsarticle}
\usepackage{todonotes}
\usepackage{xpatch}
\usepackage{array}
\usepackage{siunitx}
\usepackage{amsmath}
\usepackage{amssymb}
\usepackage{xcolor}
\usepackage{bbm}
\usepackage{amsgen,amsmath,amstext,amsbsy,amsopn,tikz,amssymb}
\usepackage{listings}
\usepackage{xcolor}
\usepackage{algorithm}
\usepackage{algorithmicx}
\usepackage{hyperref}
\usepackage[noend]{algpseudocode}
\usepackage{tabularx}
\usepackage{rotating}
\usepackage{multirow}
\usepackage{amsmath, amssymb}
\usepackage[export]{adjustbox}
\definecolor{codegreen}{rgb}{0,0.6,0}
\definecolor{codegray}{rgb}{0.5,0.5,0.5}
\definecolor{codepurple}{rgb}{0.58,0,0.82}
\definecolor{backcolour}{rgb}{0.95,0.95,0.92}
\usepackage{bm}
\usepackage{verbatim}
\usepackage{subcaption}
\usepackage[left=2.5cm, right=2.5cm, top=4cm, bottom=3cm, footskip=0.5cm]{geometry}
\usepackage{enumitem} 
\usepackage{tablefootnote}

\begin{document}

\begin{frontmatter}

\title{Bayesian Deep Learning for Discrete Choice}
\date{ }

\author[1,2]{Daniel F. Villarraga}
\ead{df.villarraga2191@uniandes.edu.co, dv275@cornell.edu}
\author[1]{Ricardo A. Daziano}
\ead{daziano@cornell.edu}

\address[1]{School of Civil and Environmental Engineering, Cornell University, 220 Hollister Hall, Ithaca, NY 14853, USA}
\address[2]{Industrial Engineering Department, Universidad de Los Andes, Carrera 1 N° 18A-12, 111711, Bogotá, Colombia}

\begin{abstract}
Discrete choice models (DCMs) are used to analyze individual decision-making in contexts such as transportation choices, political elections, and consumer preferences. DCMs play a central role in applied econometrics by enabling inference on key economic variables, such as marginal rates of substitution, rather than focusing solely on predicting choices on new unlabeled data. However, while traditional DCMs offer high interpretability and support for point and interval estimation of economic quantities, these models often underperform in predictive tasks compared to deep learning (DL) models. Despite their predictive advantages, DL models remain largely underutilized in discrete choice due to concerns about their lack of interpretability, unstable parameter estimates, and the absence of established methods for uncertainty quantification. Here, we introduce a deep learning model architecture specifically designed to integrate with approximate Bayesian inference methods, such as Stochastic Gradient Langevin Dynamics (SGLD). Our proposed model collapses to behaviorally informed hypotheses when data is limited, mitigating overfitting and instability in underspecified settings while retaining the flexibility to capture complex nonlinear relationships when sufficient data is available. We demonstrate our approach using SGLD through a Monte Carlo simulation study, evaluating both predictive metrics—such as out-of-sample balanced accuracy—and inferential metrics—such as empirical coverage for marginal rates of substitution interval estimates. Additionally, we present results from two empirical case studies: one using revealed mode choice data in NYC, and the other based on the widely used Swiss train choice stated preference data.
\end{abstract}


\end{frontmatter}

\section{Introduction}

Discrete choice is a fundamental area of econometrics that examines how individuals make decisions among a finite set of alternatives. Unlike purely predictive approaches, discrete choice analysis aims to infer the underlying decision-making process, enabling researchers to uncover behavioral patterns of preferences rather than merely forecast future choices. For example, in transportation systems, discrete choice models are often used to estimate individuals’ willingness to pay for a reduction in travel time, considering factors such as cost, trip duration, level of service, and other attributes of competing transportation modes. \\ 

Given that inference is fundamental in the discrete choice field, researchers often rely on transparent and interpretable statistical binary or multinomial classification models such as logistic and probit regressions, along with their more complex variations. Traditional discrete choice models (DCMs) allow for point and interval estimation of key economic quantities, including marginal rates of substitution and odds ratios. However, because prediction is typically not the primary objective in discrete choice studies, these models often underperform in predictive tasks compared to machine learning approaches, particularly deep learning (DL) models \cite{wang2021comparing}.\\

The primary reason deep learning models are not widely applied to discrete choice problems is their inherent lack of interpretability. Even when relevant information can be extracted from these models, their estimates may produce economic insights that deviate from established economic theory or behavioral intuition, and methods for deriving interval estimates have not been well demonstrated—concerns that have discouraged the research community. This apparent lack of interpretability arises from three key challenges: (i) model architectures that fail to incorporate expert knowledge, (ii) unstable parameter estimates, and (iii) the absence of demonstrated methods for uncertainty representation.\\

The first challenge, highlighted by \cite{wang2020deep}, involves incorporating behavioral assumptions into model design to extract economic information comparable to standard discrete choice models. The second challenge, discussed by \cite{wilson2020bayesian}, arises from under-specification, where multiple parameter estimates yield similar predictive performance but differing economic implications. The third challenge, closely related to the second, stems from limited data availability, leading to uncertainty in model-derived hypotheses \cite{wilson2020case}. To our knowledge, the limited number of deep learning applications to discrete choice have not addressed uncertainty representation and have focused solely on point estimation. Consequently, interval estimation and hypothesis testing using deep learning models in discrete choice remain open problems.\\

Deep learning models have been increasingly explored—albeit still to a limited extent—in discrete choice studies focused on point estimation and predictive performance. In such applications, DL models often achieve higher predictive accuracy (e.g., \cite{wang2021comparing}, \cite{han2022neural}, \cite{lu2021modeling}) and offer greater flexibility than traditional discrete choice models (DCMs). Additionally, previous work has demonstrated that DL models can yield economic insights as comprehensive as those derived from traditional DCMs \cite{wang2020deep}.\\

Still, DL models applied to discrete choice are often highly underspecified by the data they are trained on—typically datasets with only thousands of observations—making them prone to overfitting and resulting in unstable parameter estimates, which, in turn, affects their interpretability. To address these challenges, researchers have proposed regularization techniques and inductive biases. For instance, in \cite{wang2020deepar}, Wang introduced a novel model design that enforces independence from irrelevant alternatives (IIA) by using shared-parameter model blocks, thereby reducing model complexity. As an example of regularization, in \cite{feng2024deep}, Feng et al. proposed a gradient-based regularization technique that encourages behavioral consistency by enforcing expected gradient signs with respect to input attributes.\\

In \cite{sifringer2018enhancing}, Sifringer et al. presented a deep learning (DL) model architecture comprised of a traditional discrete choice model and a fully connected neural network. In their approach, the neural network and the traditional discrete choice model components did not share input features to ensure that the linear terms were not overrun by the nonlinearities in the neural network. Their model specification provided the first approximation of a deep learning model for discrete choice with high interpretability while achieving better predictive performance than traditional approaches. They also established a connection between multinomial logit models from discrete choice and one-dimensional convolutions from deep learning. Nevertheless, they did not address issues related to parameter instability and uncertainty representation.\\

So far, few studies have focused on uncertainty representation, interval estimation, or hypothesis testing in deep learning for discrete choice. However, in \cite{2025arXiv250309786V}, we briefly demonstrate how to construct credible intervals at the individual level using approximate Bayesian inference, attaining results that align with behavioral intuition.\\

Arguably, when addressing the issue of uncertainty representation in deep learning, the set of tools that come to mind originates from the relatively new framework of conformal prediction. In conformal prediction, uncertainty is quantified and used to create regression bands or classification sets with strong finite-sample coverage guarantees \cite{angelopoulos2021gentle}. One of its main advantages is that, in general, it requires no more than a single additional forward pass of the model through the dataset and does not depend on any model or data distribution assumptions to work as intended \cite{angelopoulos2024theoretical}. However, as its name implies, it provides a simple—yet effective—method for uncertainty quantification in prediction sets, but not for inference on parameters associated with the input variables.\\

Approximate Bayesian inference methods, on the other hand, can also be used for uncertainty representation \cite{wilson2020bayesian}, both in prediction sets and for inference on parameters associated with input variables. Bayesian Deep Learning (BDL) enables the incorporation of prior beliefs, providing a principled way to integrate expert knowledge into the modeling process—an approach that would significantly enhance efforts to apply deep learning to discrete choice. Furthermore, unlike standard deep learning, BDL aims to provide a distribution over model parameters rather than point estimates \cite{papamarkou2024position}, which can be leveraged to construct interval estimates for economic quantities such as marginal rates of substitution. \\

Multiple methods have been developed in BDL to approximate posterior distributions, including Laplace and variational approximations, ensemble methods, and posterior sampling algorithms \cite{papamarkou2024position}. For instance, Maddox et al. \cite{maddox2019simple} proposed Stochastic Weight Averaging Gaussian (SWAG), a method that captures uncertainty in weight space with minimal computational and memory overhead compared to stochastic gradient descent. However, it requires sampling from a high-dimensional multivariate Gaussian distribution to obtain interval prediction estimates.\\

Other approaches based on stochastic gradient descent do not require resampling from a high-dimensional multivariate Gaussian, as in SWAG. These methods, collectively known as stochastic gradient MCMC (SG-MCMC), are algorithms that generate posterior draws directly from the training process. In this paper, we focus on stochastic gradient Langevin dynamics (SGLD) \cite{welling2011bayesian}. While its implementation is straightforward and widely adopted in the machine learning community, we demonstrate its effective integration and adaptation to the unique challenges of the discrete choice field. Building on this foundation, we present a simple yet general DL model architecture specifically designed to leverage the power of approximate Bayesian inference methods such as SWAG and SGLD. Our goal is to develop a DL model that collapses to behaviorally informed hypotheses when data is limited, thereby avoiding overfitting and instability in underspecified settings while remaining capable of capturing complex nonlinear relationships when sufficient data is available. We demonstrate our architecture under SGLD through a Monte Carlo simulation study, which we use to estimate predictive metrics such as out-of-sample balanced accuracy and inferential metrics such as empirical coverage for marginal rates of substitution interval estimates. In addition, we provide inference results for the stated and revealed preference data, focusing on estimates for the value of travel time savings.\\

\section{Standard Discrete Choice Modeling \label{DC}}
In discrete choice theory, decision-making is modeled as a process in which an individual $i$ selects between a discrete set of alternatives (e.g., public transit or a private car for commuting) based on the maximization of their utility $u_i$ (unobserved to the modeler). This utility consists of a deterministic component $v_i$, which accounts for observable characteristics of the individual and the alternatives, as well as a stochastic component that captures unobserved factors.\\

In a standard binary discrete choice framework, the utility $u_i$ associated with selecting one alternative over another (e.g., choosing transit over driving) is expressed as:
\begin{equation}
u_i = v_i + \epsilon_i,
\label{eq:Utility}
\end{equation}
where $\epsilon_i$ represents the unobserved component of utility, and $v_i$ (often referred to as the representative part of the utility) is usually defined as the following linear function:
\begin{equation}
v_i = \bm{x_i}^T \bm{\beta} + \bm{q_i}^T \bm{\gamma}.
\label{eq:Vut}
\end{equation}
Here, $\bm{\beta}$ and $\bm{\gamma}$ are estimable parameter vectors, $\bm{x_i}$ captures differences in mode attributes (e.g., travel time differences), and $\bm{q_i}$ represents individual-specific characteristics (e.g., income level). The form of the model--rarely nonlinear in attributes and parameters, including interactions between parameters and explanatory variables, is typically guided by economic theory and domain expertise.\\

The specific distribution of $\epsilon_i$ determines the model type: a standard normal assumption leads to a binary probit model, whereas a standard logistic distribution corresponds to logistic regression or a binary logit model\footnote{In the binary logit model, the utility functions include i.i.d. extreme value type 1 (EV1) error terms. The likelihood depends on the difference in utilities, which results in the utility difference $u_i$ as defined above. Due to the properties of the EV1 distribution, the difference in error terms follows a logistic distribution, making the model specification mathematically equivalent to logistic regression.}.\\

The probability of selecting the preferred alternative (i.e., $y_i = 1$, where $y_i$ is a binary choice indicator) can be expressed as:
\begin{equation}
P(y_i = 1) = P(\epsilon_i \leq v_i),
\label{eq:pstandard}
\end{equation}
and for a sample of $n$ individuals, the likelihood function of this model is given by:
\begin{equation}
\mathcal{L}(\bm{\beta}, \bm{\gamma}) = \prod_{i=1}^n P(\epsilon_i \leq v_i)^{y_i} \left(1-P(\epsilon_i \leq v_i)\right)^{1-y_i}.
\label{eq:lstandard}
\end{equation}

It is evident that this binary specification is equivalent to logistic regression, and the log-likelihood function corresponds to the binary cross-entropy loss function.\\

\begin{figure}[h]
    \centering
    \includegraphics[width=0.7\linewidth]{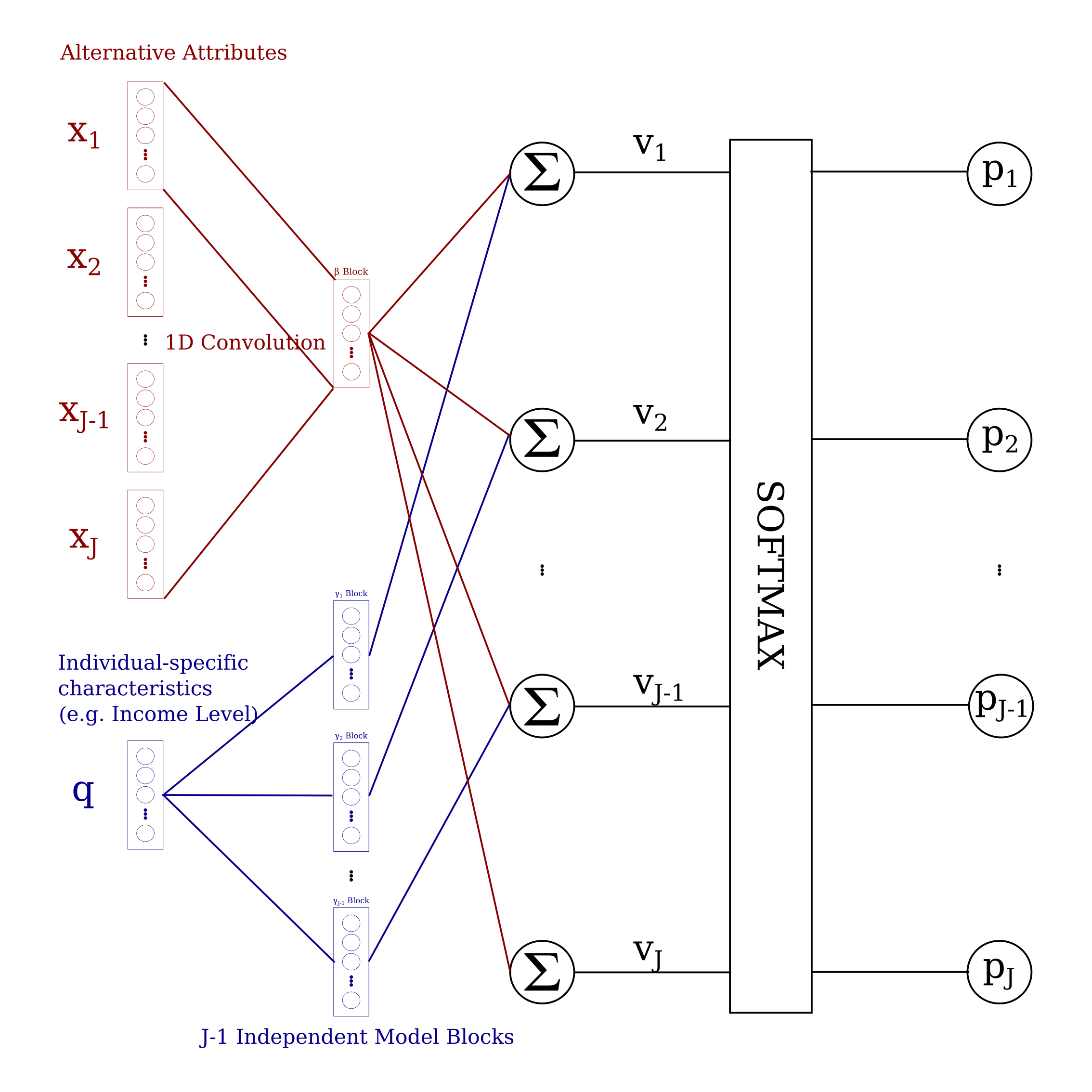}
    \caption{Conditional logistic regression as a shallow neural network.}
    \label{fig:multinomial_shallow}
\end{figure}

In the multinomial setting, a typical specification consists of latent utility functions for each alternative $j$, expressed as:
\begin{equation}
    u_{ij} = \bm{x_{ij}}^T \bm{\beta} + \bm{q_i}^T \bm{\gamma_j} + \epsilon_{ij},
\end{equation}
where the alternative attribute parameters $\bm{\beta}$ are shared across alternatives, while $\bm{\gamma_j}$ are defined only for $J-1$ alternatives to ensure identifiability. A representation of this model as a shallow neural network is provided in Figure~\ref{fig:multinomial_shallow}. As depicted in the figure, the alternative-specific attributes undergo a 1D convolution with a kernel size equal to the number of alternative attributes, while the shared individual-specific characteristics pass through independent model blocks with parameters equal to the number of individual-specific characteristics. The relevant outputs from the 1D convolution for each alternative, along with those from the $J-1$ independent model blocks, are summed per alternative to obtain $J$ representative utility values.\footnote{For alternative $J$, the representative utility depends only on the relevant output from the 1D convolution. Alternative $J$ serves as the base alternative, as all alternative parameters related to individual-specific attributes are computed relative to it, ensuring model identifiability.} These representative utilities then pass through a Softmax activation function to generate probability predictions for each alternative, representing the likelihood of the individual choosing each option given their characteristics and the attributes of the alternatives. \\

For a more detailed and general description of standard discrete choice models, refer to \cite{train_2009}.\\

\section{Deep Learning for Discrete Choice}

In discrete choice research, inference is often considered more important than prediction. Researchers and practitioners are generally more interested in understanding the underlying data-generating process than in achieving high predictive performance without economic insights. Consequently, the inherent lack of interpretability in deep learning models and their tendency to overfit the training set when data is limited have discouraged the discrete choice community from adopting these models. \\

The lack of interpretability in deep learning models applied to discrete choice stems from three main challenges, namely: (i) model architectures that lack behavioral interpretation, (ii) unstable parameter estimates, and (iii) a lack of methods for representing epistemic (and aleatoric) uncertainty.\\

The first challenge was highlighted by \cite{wang2020deep}, who demonstrated how to extract economic information as comprehensive as that provided by standard discrete choice models and illustrated methods for incorporating behavioral assumptions into model specifications through architectural design. \\

The second challenge concerns the shape of the cost function during training when the model is underspecified by the data, as discussed by \cite{wilson2020bayesian}. Specifically, when the model architecture includes a large number of parameters but is constrained by limited data, the cost function tends to exhibit extensive valleys and multiple modes, leading to numerous local minima. This complex landscape can yield different parameter estimates with equivalent predictive performance in the training data but distinct and sometimes implausible economic interpretations.\\

The third challenge—closely related to the second—is that, in most discrete choice inference problems, researchers work with very limited data, leading to uncertainty in the hypotheses derived from the model-fitting process \cite{wilson2020case}. To our knowledge, only a limited number of deep learning applications to discrete choice have addressed the representation of epistemic uncertainty, as most have primarily focused on point estimation.\\

For instance, very recently, Zheng et al. \cite{zheng2024incorporating} used Bayesian Neural Networks (BNNs) to quantify prediction uncertainty for guiding more efficient survey data collection in travel mode choice. However, they did not propose an interpretable model that could be used to derive economic insights, nor did they compute point or interval estimates for marginal utilities or other relevant information. Nevertheless, their application of BNNs demonstrates how uncertainty quantification can be leveraged in discrete choice settings, even without interpretable models. \\

Arguably, the primary focus in the field of deep learning for discrete choice has been on designing architectures that incorporate economic assumptions and expert knowledge, leading to constrained model classes with better generalization. Some studies have proposed architectures that enforce independence from irrelevant alternatives by structuring neural networks with independent model blocks for each alternative, resulting in architectures similar--although deeper--to the one presented in Figure \ref{fig:multinomial_shallow} [e.g., \cite{wang2020deep}, \cite{arkoudi2023combining}, \cite{2025arXiv250309786V}].\\

Other works have focused on augmenting discrete choice models with deep learning architectures while preserving their interpretability. The quintessential example of this approach comes from Sifringer et al. \cite{sifringer2018enhancing}, who formulate a deep learning discrete choice model with two distinct components: an interpretable model informed by expert knowledge, which depends on attributes relevant for inference, and a data-driven deep learning model with input attributes that, while not relevant for inference, may still inform individual utilities.\\

Following a similar approach, Wong et al. \cite{wong2021reslogit} propose a Residual Neural Network (ResLogit), in which a representative portion of the utilities depends on alternative-specific attributes, while another portion is influenced by attributes from all alternatives. This structure allows the model to learn complex substitution patterns in an interpretable manner. However, their model does not explicitly address how to incorporate attributes shared by all alternatives, such as individual characteristics. Additionally, the linear component of the utility may be overshadowed by the nonlinear component that depends on all alternative attributes.\\

\begin{figure}[h!]
\begin{minipage}[c]{0.495\linewidth}
    \centering
    \includegraphics[width=\linewidth]{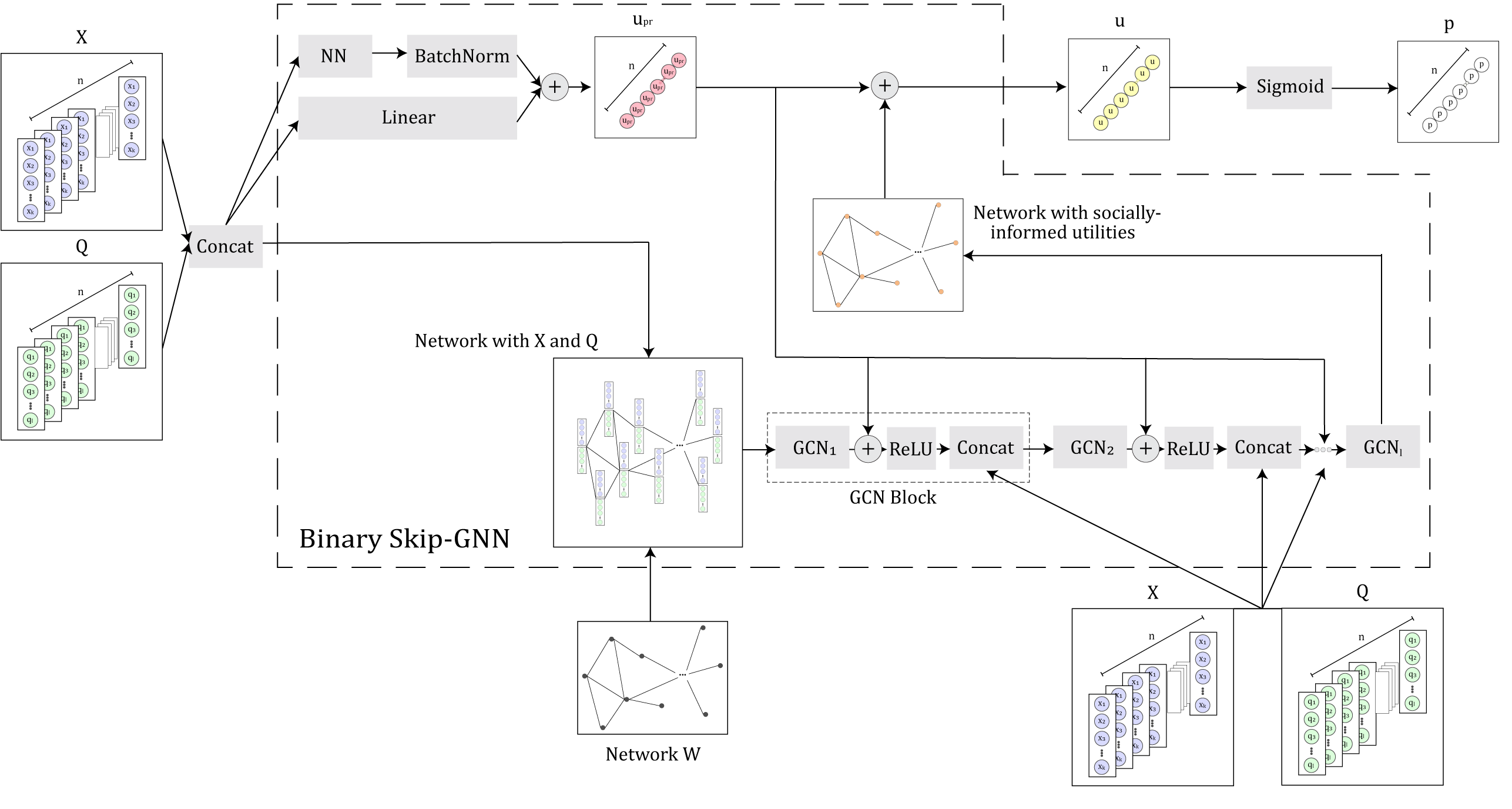}
    \caption{Binary Skip-GNN model architecture\cite{2025arXiv250309786V}.}
    \label{fig:skip_gnn_binary}
\end{minipage}
\hfill
\begin{minipage}[c]{0.495\linewidth}
    \centering
    \includegraphics[width=\linewidth]{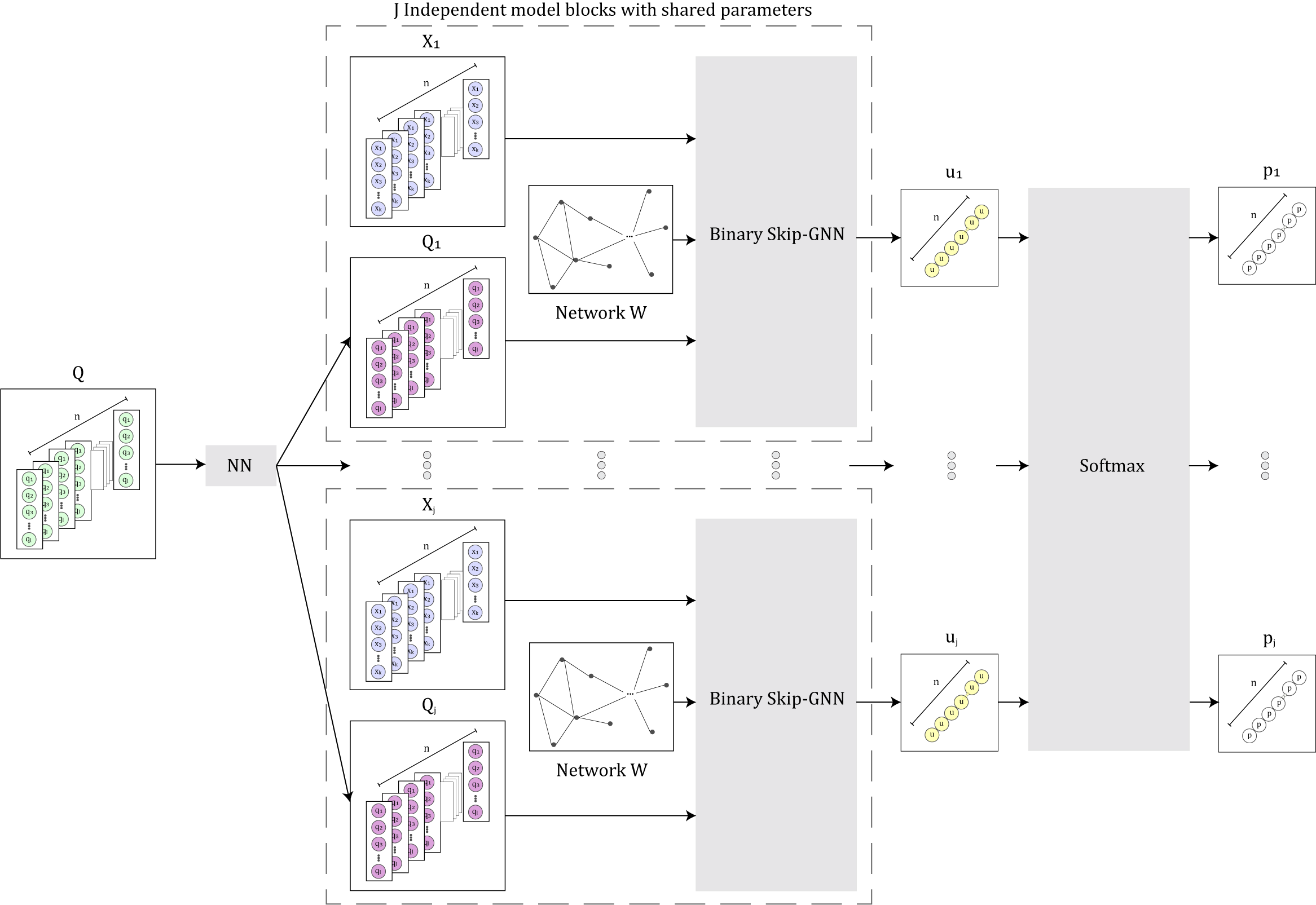}
    \caption{Multinomial Skip-GNN with IIA model architecture \cite{2025arXiv250309786V}.}
    \label{fig:skip_gnn_multi}
\end{minipage}
\end{figure}

In \cite{2025arXiv250309786V}, we introduce an architecture (Figures \ref{fig:skip_gnn_binary} and \ref{fig:skip_gnn_multi}) that models scenarios with IIA, leverages residual connections for improved identifiability, enables automatic feature learning and attribute interactions, controls non-linearity magnitudes, incorporates individual characteristics, and captures peer effects using graph convolutional neural networks. Building on these innovations, this paper proposes a general deep learning model designed to work effectively with approximate Bayesian inference.\\

\section{Approximate Bayesian Inference for Deep Learning Models}

A common concern with deep learning models is their tendency to overfit training datasets due to the neglect of epistemic and aleatoric uncertainty \cite{Wilson2020}. Their flexibility allows them to represent multiple hypotheses consistent with the training data. Some studies [e.g., \cite{Li2018}] examine the geometry of the loss surface in deep learning models across various domains. Researchers have found that these landscapes are often highly non-convex, featuring multiple local optima and large connected valleys—where different configurations of model parameters $\bm{\Theta}$ achieve nearly the same posterior probability. However, non-Bayesian deep learning approaches rely on a single model, disregarding uncertainty that could be captured by disagreement among multiple plausible models—each with high posterior probability—resulting in poor generalization.\\

A key advantage of Bayesian deep learning is the use of marginalization rather than optimization \cite{Wilson2020}. The predictive distribution in a Bayesian setting is given by Equation \ref{eq:marg2}, where different parameter sets $\bm{\Theta}$ are weighted by their posterior distribution $p(\bm{\Theta}|\bm{y},\bm{X}, \bm{Q})$. If this integral were analytically computable, predictions would follow a Bayesian Model Average (BMA) \cite{Raftery1997}. 

\begin{equation}
p(y_{ij}| \bm{y}, \bm{X}, \bm{Q}) = \int_{\bm{\Theta}} p(y_{ij}|\bm{X}, \bm{Q},\bm{\Theta}) p (\bm{\Theta} | \bm{y}, \bm{X}, \bm{Q}) d\bm{\Theta}
\label{eq:marg2}
\end{equation} 

The extent to which Bayesian and classical approaches differ depends on the posterior distribution’s shape. When the posterior is sharply concentrated around the maximum a posteriori (MAP) estimate, $\bm{\Theta}_{MAP}$, both approaches yield similar results. Moreover, as training datasets grow, epistemic uncertainty decreases, the posterior distribution converges to the MAP, and Bayesian and classical deep learning models provide comparable predictions and insights.\\

In discrete choice applications, where data availability is often limited, deep learning models tend to be underspecified. This leads to unstable parameter estimates, as multiple model configurations can yield equivalent values for the cost function or posterior distribution $p(\bm{\Theta}|\bm{y},\bm{X}, \bm{Q})$. Consequently, selecting a model parametrized by $\bm{\Theta}_1$ may produce similar predictive performance to another parametrized by $\bm{\Theta}_2$, yet lead to economic interpretations that contradict each other or deviate from behavioral intuition.\\

Recent applications of deep learning (DL) to discrete choice modeling suggest that model interpretability and generalization can be improved by leveraging ensemble methods \cite{wang2020deep}. Since deep learning architectures are often underspecified due to limited training data, they tend to exhibit irregular loss landscapes with large connected valleys and multiple modes \cite{izmailov2018averaging, li2018visualizing}. This results in diffuse posteriors with significant parameter uncertainty, making Bayesian approaches particularly valuable for these models. By marginalizing over a set of $\bm{\Theta}$ configurations instead of relying on a single set of model weights, Bayesian deep learning effectively captures uncertainty and improves robustness by averaging multiple high-performing models.  \\

To address these challenges, researchers can integrate into their studies methods such as Stochastic Weight Averaging (SWA), Stochastic Weight Averaging Gaussian (SWAG) \cite{izmailov2018averaging}, and Stochastic Gradient Langevin Dynamics (SGLD) \cite{welling2011bayesian}, which approximate the posterior while maintaining computational efficiency. Deep ensembles \cite{Lakshminarayanan2017}, another widely used technique, have been shown to serve as posterior approximations \cite{Wilson2020}, further demonstrating the benefits of Bayesian-inspired approaches. These methods not only mitigate overfitting but also facilitate uncertainty quantification for both prediction and inference. Leveraging posterior information in this way enables the construction of credible intervals for model parameters, reinforcing the case for deep learning in discrete choice applications.

\subsection{Approximation of the Posterior for Bayesian Deep Learning}

There are multiple methods to approximate the posterior distribution $p(\bm{\Theta}|\bm{y},\bm{X},\bm{Q})$ in Bayesian deep learning \cite{papamarkou2024position}. Some of these methods include, but are not limited to posterior sampling [e.g. \cite{welling2011bayesian}] and Laplace and variational approximations [e.g. \cite{10.1162/neco.1992.4.3.415}]. Ultimately, the primary goal is to approximate or sample from the posterior distribution in regions of high density while capturing a set of parameters that induce an informative distribution--that captures epistemic and aleatoric uncertainty--over function predictions, $f(\bm{X}, \bm{Q};\bm{\Theta})$. In discrete choice, this function corresponds to the latent representative utility $\bm{v}$, as defined in previous sections. \\

\begin{figure}[h!]
    \centering
    \includegraphics[width=0.5\linewidth]{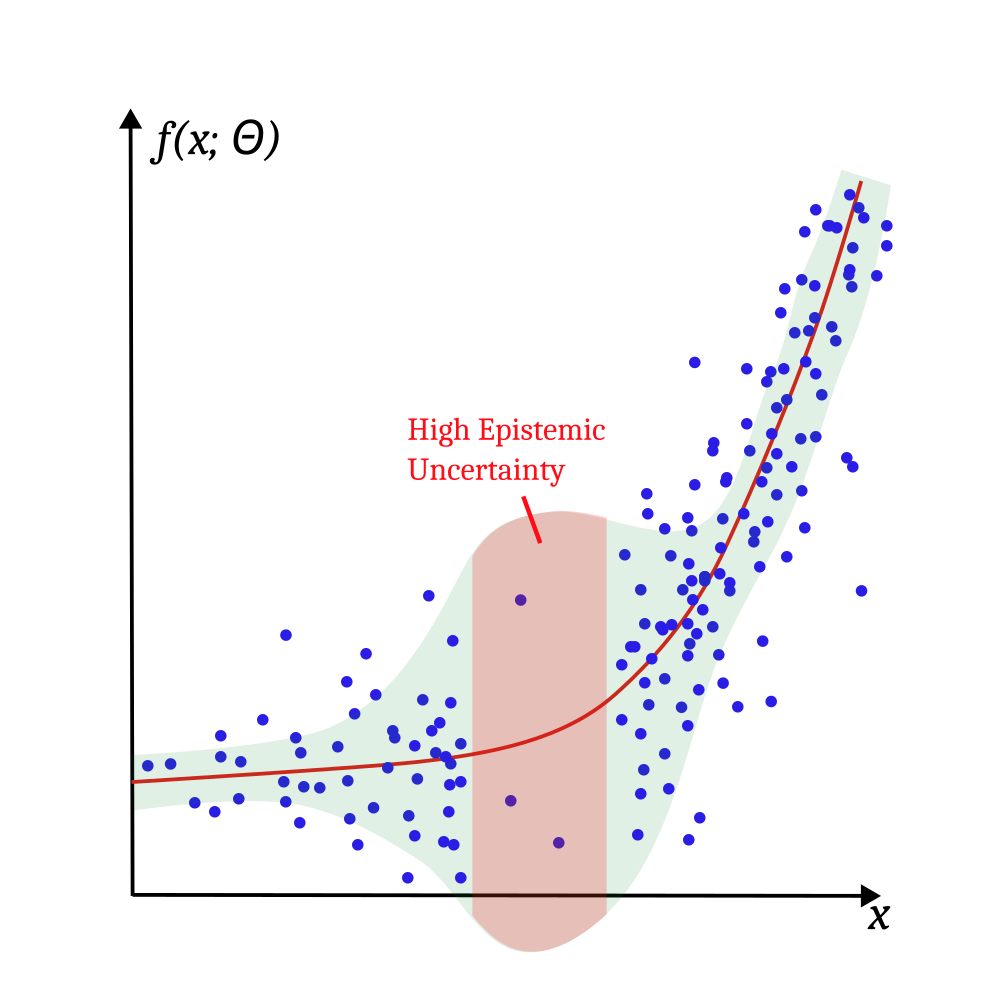}
    \caption{Representation of uncertainty in the prediction $f(\bm{x};\bm{\Theta})$. The prediction bands represent $(1-\alpha)$ credible intervals. The solid red line depicts the true data-generating process, while observations are represented by blue dots. The highlighted region shows an interval in the input space with high epistemic uncertainty.}
    \label{fig:e_uncertainty}
\end{figure}

In other words, if a finite number of parameter sets $\bm{\Theta}$ are used to approximate the posterior, each $\bm{\Theta}$ should have a high posterior probability while collectively capturing epistemic and aleatoric uncertainty. This is achieved by having estimates of $f(\bm{X}, \bm{Q};\bm{\Theta})$ that vary in regions of the input space with scarce observations and converge in regions where data is abundant \cite{Wilson2020} and aleatoric uncertainty is low. A good approximation to the posterior distribution will ensure a good representation of uncertainty, which is illustrated in figure \ref{fig:e_uncertainty}\\

\subsubsection{Deep Ensembles}

Deep ensembles are models that have empirically demonstrated improvements in accuracy over single deep learning models \cite{Lakshminarayanan2017}. A deep ensemble is essentially a collection of deep learning models that are used collectively to make predictions. In the past, deep ensembles have been considered alternatives to Bayesian deep learning; however, Wilson and Izmailov argued that they are simply another way to approximate the posterior \cite{Wilson2020}.\\

Deep ensembles can be trained in parallel using a single dataset and model architecture. Lakshminarayanan et al. \cite{Lakshminarayanan2017} developed an adversarial algorithm to train each model independently. However, the adversarial component can be omitted, and a deep ensemble can be constructed by optimizing the posterior density multiple times from different random initializations in the weight space. They propose that the predictions of each model in the ensemble are weighted equally—equivalently assuming that each weight configuration has the same a posteriori probability. Consequently, the predictions from a deep ensemble are given by:

\begin{equation}
p(\bm{y}|\bm{X}) = M^{-1} \sum_{m=1}^M p(\bm{y}|\bm{X}, \bm{\Theta}_m)
\end{equation}

where each $\bm{\Theta}_m$ is a local maximizer of the posterior $p(\bm{\Theta}|\bm{y},\bm{X})$. This approach performs well in cases where the aleatoric initialization in weight space ensures convergence to multiple modes in the posterior and where the modes of the posterior to which it converges are sharp and have equivalent densities. However, random initializations may converge to the same mode, the modes of the posterior may have significantly different posterior probabilities, and competing models within each posterior mode may be discarded in favor of local maximizers with small gains in posterior probability. These issues can lead to a poor representation of uncertainty.

\subsubsection{Stochastic Weight Average (SWA) and SWA-Gaussian (SWAG)}
Stochastic Weight Averaging (SWA) is a computationally efficient way to deal with unstable parameter estimates. SWA offers better generalization and more stable solutions than those obtained from standard training methods while potentially providing better behavioral insights. The SWA approach averages the parameter iterates from stochastic gradient descent during the learning process. The algorithm introduces minimal computational and memory overhead since the parameter average is computed as a running average every \(c\) gradient updates. A simplified version of the learning procedure, presented in \cite{izmailov2018averaging}, is shown below:

\begin{algorithm}
\caption{Stochastic Weight Averaging \cite{izmailov2018averaging}}
\begin{algorithmic}[1]
\Require initial parameters \(\tilde{\bm{\Theta}}\), cycle length \(c\), number of epochs \(e\), learning rate \(\alpha\), loss function \(L\)
\State \(\bm{\Theta} \gets \tilde{\bm{\Theta}}\) \Comment{Initialize parameters with \(\tilde{\bm{\Theta}}\)}
\State \(\bm{\Theta}_{SWA} \gets \tilde{\bm{\Theta}}\) \Comment{Initialize SWA parameters}
\State \(\eta_{models} \gets 0\) \Comment{Initialize number of models in average}
\For{\(i \gets 1, 2, \dots, e\)}
    \State \(\bm{\Theta} \gets \bm{\Theta} - \alpha \nabla L(\bm{\Theta})\) \Comment{Stochastic gradient update}
    \If{\(\mod(i, c) = 0\)}
        \State \(\bm{\Theta}_{SWA} \gets \frac{\bm{\Theta}_{SWA} \cdot \eta_{models} + \bm{\Theta}}{\eta_{models} + 1}\) \Comment{Compute model average}
        \State \(\eta_{models} \gets \eta_{models} + 1\) \Comment{Increment the number of models in the average}
    \EndIf
\EndFor
\end{algorithmic}
\end{algorithm}

SWAG has a very similar implementation to SWA but differs in that it also computes the running parameters' variance and uses these statistics to approximate the posterior with a high-dimensional multivariate normal distribution. In that sense, SWA only provides a set of parameters in a flatter region of the parameter space that has the potential for better generalization \cite{izmailov2018averaging} but remains a single point estimate, while SWAG gives a true approximation to the parameter posterior that can be used for Bayesian inference.\\

\subsubsection{Stochastic Gradient Langevin Dynamics}

Similar to SWA and SWAG, SGLD uses iterates from stochastic gradient descent. It approximates Langevin dynamics to obtain an approximation of the parameters posterior distribution that can be treated as MCMC iterates and used for Bayesian inference. The algorithm works by injecting noise \(\bm{\eta}_t\) into the (batched) stochastic gradient updates, as described by equations \ref{eq:sgld_1} and \ref{eq:sgld_2}, and saving the parameter iterates \(\bm{\Theta}_t\). In those equations, \(q_{ti}\), \(x_{ti}\), and \(y_{ti}\) represent the socio-demographics, alternative attributes, and selected alternative for individual \(i\) in batch \(t\). \(p(\bm{\Theta}_t)\) is the prior distribution for the parameters (e.g., induced by the \(\ell_1\) or \(\ell_2\) regularization in deep learning frameworks), and \(p(q_{ti}, x_{ti}, y_{ti} \mid \bm{\Theta}_t)\) is the likelihood of the observed individual under the set of parameters \(\bm{\Theta}_t\). In Equation \ref{eq:sgld_1}, \(N\) denotes the total number of training observations, \(n\) denotes the number of observations in the batch, and $\alpha_t$ denotes the gradient step size at $t$.

\begin{equation}
\Delta \bm{\Theta}_t = \frac{\alpha_t}{2} \left( \nabla \log p(\bm{\Theta}_t) + \frac{N}{n} \sum_{i=1}^n \nabla \log p(q_{ti}, x_{ti}, y_{ti} \mid \bm{\Theta}_t) \right) + \bm{\eta}_t
\label{eq:sgld_1}
\end{equation}

\begin{equation}
\bm{\eta}_t \sim \mathcal{N}(0, \alpha_t)
\label{eq:sgld_2}
\end{equation}

With this gradient update structure, \(\bm{\Theta}_t\) approaches samples from the posterior \(p(\bm{\Theta} \mid q_{ti}, x_{ti}, y_{ti})\), as these updates approximate Langevin dynamics, which converge to the posterior distribution \cite{welling2011bayesian}. In contrast with SWAG, SGLD does not approximate the posterior mode using a Gaussian distribution and does not require additional post-estimation sampling.\\

In this paper, we implement SGLD to perform approximate Bayesian inference by directly sampling the posterior distribution using our proposed deep learning architecture.

\section{A General Deep Learning Discrete Choice Model Architecture}

Our model architecture builds on some of the foundations we established in \cite{2025arXiv250309786V}. It consists of four distinct components:  

\begin{enumerate}[label=\roman*)]
    \item An embedding layer for shared input variables across alternatives (as presented in\cite{arkoudi2023combining} for better interpretability).
    \item A knowledge informed component that depends on the alternative attributes and shared individual characteristics. Usually, this component would be linear in inputs and parameters.
    \item A non-linear component that is independent across alternatives (IIA nonlinearity).
    \item A non-linear component that captures correlations across alternatives (non-IIA nonlinearity).
\end{enumerate}

\begin{figure}[h]
    \centering
    \includegraphics[width=1\linewidth]{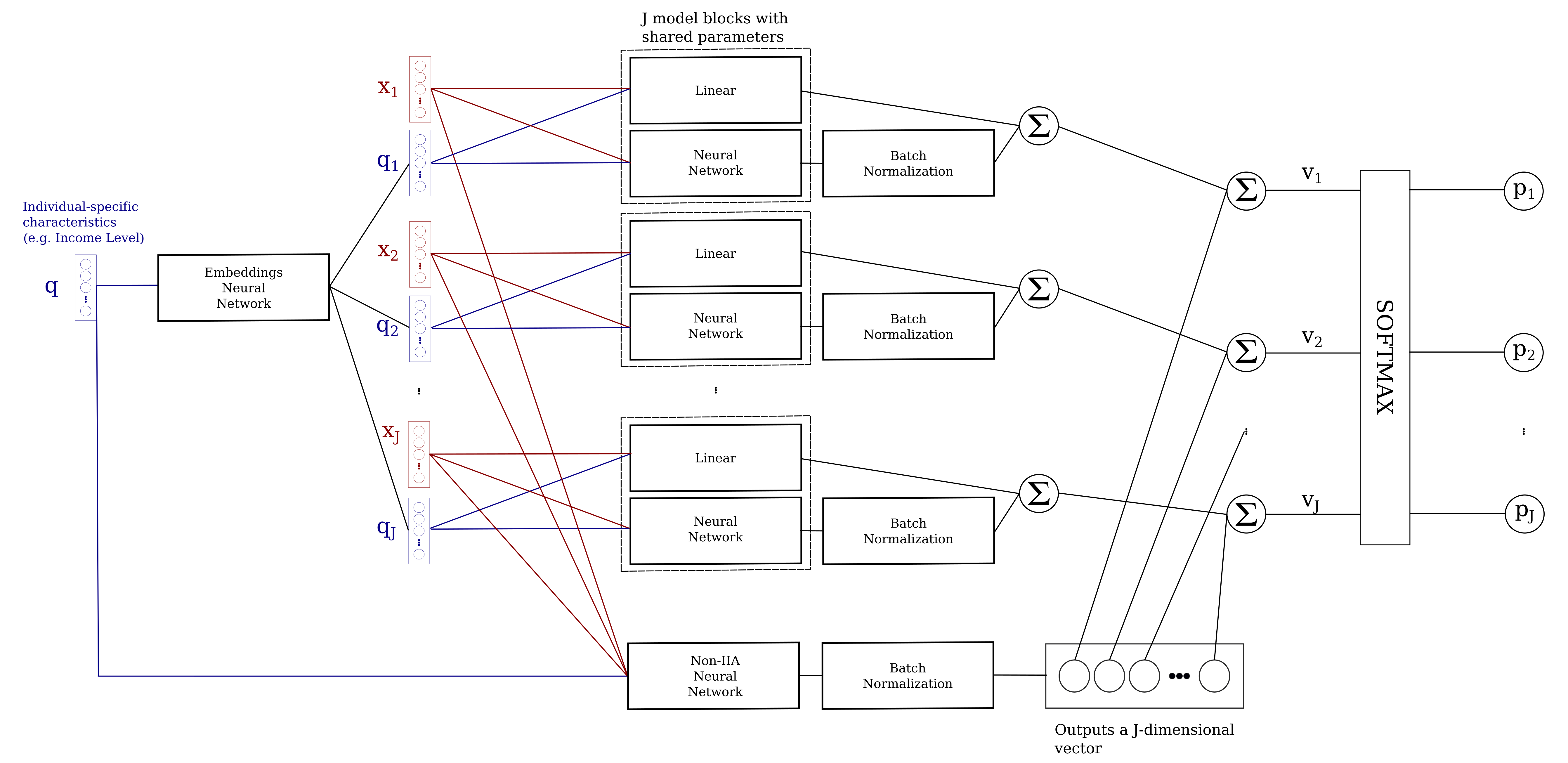}
    \caption{Proposed deep learning model architecture.}
    \label{fig:proposed_nn}
\end{figure}

A graphical representation of the proposed architecture is shown in Figure \ref{fig:proposed_nn}. As illustrated, the shared input variables across alternatives (e.g., socio-demographics such as income level), denoted by $\bm q$, pass through an embedding layer that generates $J$ embeddings of the same size as the shared input variables. These embeddings are then concatenated with the alternative-specific attributes $\bm x_j$ (e.g., trip cost or travel time).  \\

For each alternative, the alternative-specific attributes and embeddings are processed through:  
i) a knowledge informed  simple model (e.g. linear in both inputs and parameters) that could be shared across alternatives, and ii) a fully connected neural network that learns interactions between attributes and individual characteristics, with parameters that can also be shared across alternatives.  \\

The output from the neural network is passed through a batch normalization layer and summed with the output from the knowledge informed model, forming the non-linear IIA block of the model. \\ 

The non-linear, non-IIA block consists of a neural network that takes as input the original socio-demographics concatenated with all alternative attributes. The resulting $J$-dimensional output undergoes batch normalization without learnable parameters--i.e. not affine BatchNorm. The outputs from both the non-linear IIA block and the non-IIA block are then summed to compute the representative utility for each alternative. Finally, the representative utilities are passed through a Softmax activation function to obtain the choice probabilities for each alternative.\\

The representative utilities in our model can be expressed as:

\begin{equation}
    \bm{v}_{ij} = \alpha_j + \bm{x}_{ij}^{\top} \bm{\beta} + \bm{q}_{ij}^{\top} \bm{\gamma} + \sigma_{IIA}\times\text{BatchNorm}[f_{\Theta_f}(\bm{x}_{ij}, \bm{q}_{ij})] + \sigma_{nonIIA}\times\text{BatchNorm}[g_{\Theta_g}(\bm{x}_{i}, \bm{q}_{i})_j],
\end{equation}

where $\bm{q}_{ij}$ is the embedding representation for alternative $j$ and individual characteristics $\bm{q}_i$. The BatchNorm layers are not affine, meaning that the output is not rescaled and relocated. Instead, parameters $\sigma_{IIA}$ and $\sigma_{nonIIA}$ are introduced to scale the outputs from those components, but we purposely exclude the relocation parameters as the location of the latent utilities $v_{ij}$ with respect to the base alternative is already determined by the alternative specific constant $\alpha_j$.  The BatchNorm layers also ensure that the nonlinear component of our model does not overtake the knowledge-informed part of the model, which may be as simple as being linear in attributes and parameters.\\

This architecture resembles that proposed by Wang et al.~\cite{wang2021theory} in that it combines a knowledge-informed component with a deep learning model capable of automatic feature learning. However, their architecture (i) does not fix the location of the utilities, (ii) does not include separate IIA and non-IIA nonlinear components, and (iii) introduces a weighting parameter $\lambda$ for the linear combination of components, but does not relate it directly to their relative scale, since the components are not explicitly scaled (which could hinder convergence). Our model incorporates these three points.\\

The linear component of the model, $\alpha_j + \bm{x}_{ij} \bm{\beta} + \bm{q}_{ij} \bm{\gamma}$, is used to represent the part of the model informed by expert knowledge and behavioral assumptions. However, in practice, this component can have any other form and include parameter interactions and non-linear parameterized components. Creating embeddings for $\bm{q}_{ij}$ is necessary to ensure identifiability of the linear model, similar to the approach proposed in \cite{arkoudi2023combining}. The main difference is that we allow individual embeddings to interact with alternative-specific attributes in the non-linear IIA block. However, alternatively, the informed component could be formulated with $\bm{\gamma}_J = \bm{0}$, or adapted to incorporate other assumptions relevant to each case study.\\ 

The non-linear component consists of an IIA block, $\text{BatchNorm}[f_{\Theta_f}(\bm{x}_{ij}, \bm{q}_{ij})]$ parametrized by $\Theta_f$, and a non-IIA block, $\text{BatchNorm}[g_{\Theta_g}(\bm{x}_{i}, \bm{q}_{i})_j]$ parametrized by $\Theta_g$. Batch normalization stabilizes the model parameters by computing batch statistics to normalize outputs across the batch, preventing internal covariate shift \cite{ioffe2015batch}. Additionally, batch normalization has been shown to accelerate training and improve performance \cite{ioffe2015batch}. In this model, the BatchNorm layers do not have any learnable parameters, and the outputs are rescaled (by $\sigma_{IIA}$ or $\sigma_{nonIIA}$) but not shifted. Its inclusion in our neural network can be interpreted as a mechanism to determine the general location of the representative utilities while providing a straightforward--interpretable--way to impose priors on the scale of the non-linear components (by penalizing the scales $\sigma_{IIA}$ or $\sigma_{nonIIA}$ instead of relying on standard $\ell_2$ or $\ell_1$ regularization, which penalizes the norms of all neural network weights).\\ 

Our model's structure allows it to capture interactions between input variables and complex substitution patterns while also collapsing to simpler hypotheses and maintaining independence from irrelevant alternatives (when more complex hypotheses cannot be supported by the data under the predefined priors), particularly under a Bayesian framework with appropriate parameter priors applied to specific model blocks by controlling the scales $\sigma_{IIA}$ and $\sigma_{nonIIA}$. For instance, consider the following loss function that resembles a cross-entropy loss with $\ell_2$ regularization:
\begin{equation}
    \mathcal{L} = -\sum_{i=1}^{N} \sum_{j=1}^{J} y_{ij}\log p(y_i = j \mid \bm{\beta}, \bm{\gamma}, \bm{x}_i, \bm{q}_i, \bm{\Theta_f}, \bm{\Theta_g}) + \frac{\lambda_{IIA}}{2} \|\sigma_{IIA}\|_2^2 + \frac{\lambda_{nonIIA}}{2} \|\sigma_{nonIIA}\|_2^2
\end{equation}
where:
\begin{itemize}
    \item $y_{ij}$ takes a value of one if alternative $j$ was chosen by individual $i$ and zero otherwise.
    \item $\lambda_{IIA}$ and $\lambda_{nonIIA}$ are the regularization hyperparameters that induce priors on the scale of the nonlinear components in the model.
    \item $\|\sigma\|_2^2$ represents the squared $\ell_2$-norm of the scale parameters.
\end{itemize}

From a Bayesian perspective, this loss function is equivalent to a negative log-posterior, where the log-likelihood corresponds to the double summation, and the prior is determined by the $\ell_2$ regularization terms. Here, the prior follows a multivariate Gaussian distribution with a mean of zero and variance proportional to $\lambda_{IIA}^{-1}$ and $\lambda_{nonIIA}^{-1}$ for $\sigma_{IIA}$ and $\sigma_{nonIIA}$, and infinite variance for $\bm{\beta}$ and $\bm{\gamma}$—resulting in an uninformative prior for these two parameter vectors. \\

This formulation allows the model to represent assumptions about the strength of non-linearities and complex substitution patterns, ensuring that it collapses to the expert-informed component when insufficient data is available to support more complex hypotheses. From this model, it is straightforward to see that the effect of the $k$-th alternative-specific attribute $(x_{ij})_k$ on the latent utility for alternative $j$ and individual $i$ can be computed as:
\begin{equation}
    \frac{\partial u_{ij}}{\partial (x_{ij})_k} = \beta_k + \frac{\partial f_{\Theta_f}(\bm x_{ij}, \bm q_{ij})}{\partial [\bm x_{ij}]_k} \times \sigma_{IIA} + \frac{\partial g_{\Theta_g}(\bm x_{i}, \bm q_i)_j}{\partial [\bm x_{ij}]_k} \times {\sigma_{nonIIA}}
\end{equation}
which consists of the linear effect on the utilities, the non-linear but IIA effect which scale depends on $\sigma_{IIA}$, and the non-linear non-IIA effect on the latent utilities with scale determined by $\sigma_{nonIIA}$. To compute the marginal rate of substitution between attributes $k$ and $l$, one can take the ratio:
\begin{equation}
    \frac{\partial [\bm x_{ij}]_k}{\partial [\bm x_{ij}]_l} = \frac{\partial u_{ij}}{\partial (x_{ij})_l} \times \left(\frac{\partial u_{ij}}{\partial (x_{ij})_k}\right)^{-1}
\end{equation}
Note that when the non-linear effects are negligible, either due to strong priors or other factors, inference with the full model is equivalent to the simpler model specification $v_{ij} = \bm{x}_{ij} \bm{\beta} + \bm{q}_{ij} \bm{\gamma}$, so that marginal rates of substitution are given by the fixed parameter ratios $\beta_l / \beta_k$, which are the ones from the knowledge-informed model.\\

\subsection{Interpretability and Identifiability}
\begin{figure}[t]
    \centering
    \begin{minipage}[t]{0.48\linewidth}
        \centering
        \includegraphics[width=\linewidth]{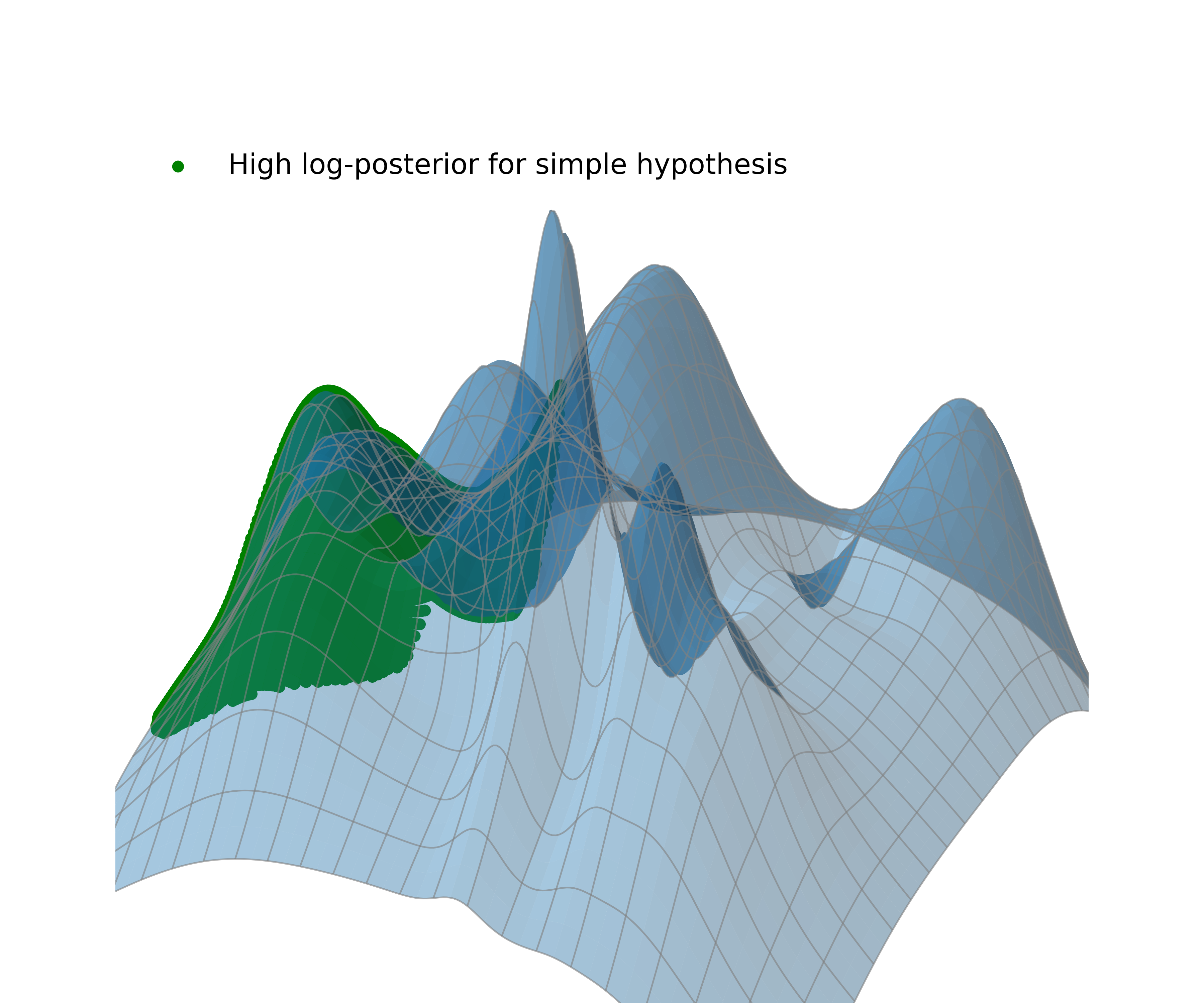}
        \caption{Step 1: Convergence toward a region (green) of the parameter space where simple hypotheses attain large log-posterior values.}
        \label{fig:step1}
    \end{minipage}\hfill
    \begin{minipage}[t]{0.48\linewidth}
        \centering
        \includegraphics[width=\linewidth]{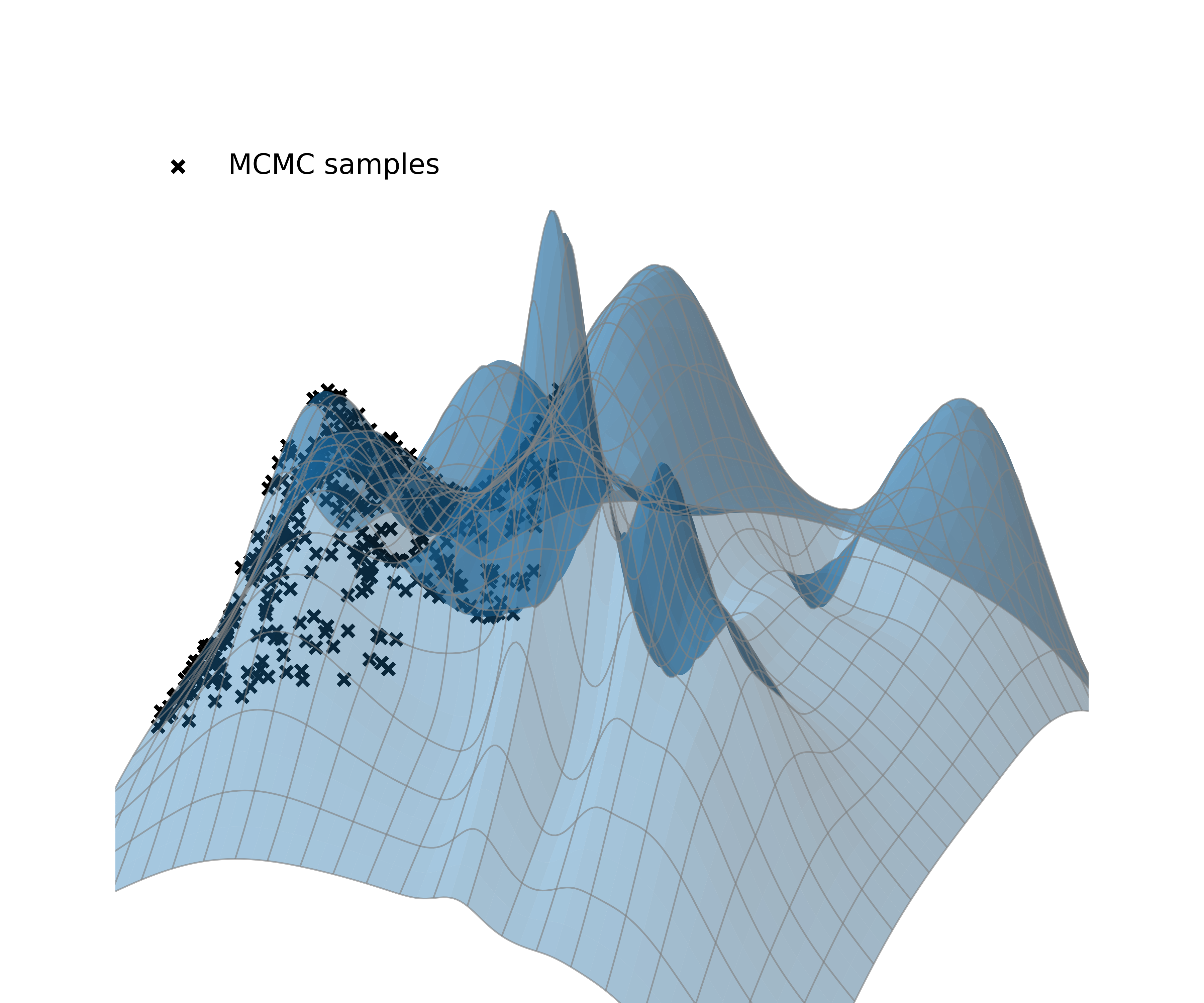}
        \caption{Step 2: Sampling (black dots) close to a convergence mode within a region where simple hypotheses attain large log-posterior values.}
        \label{fig:step2}
    \end{minipage}
\end{figure}

It is well known that deep learning models are not identifiable. However, the structure of our model, coupled with a two-step learning procedure, stabilizes parameter estimates around a convergence mode in a region of the parameter space where simple hypotheses attain large log-posterior values (relative to other regions). Our proposed two-step learning process is depicted in Figures \ref{fig:step1} and \ref{fig:step2}.\\

Our learning procedure ensures that the model defaults to the behaviorally informed hypothesis when the data do not support more complex explanations. First, we optimize the objective (log-posterior) while keeping the nonlinear layers frozen until convergence (Figure \ref{fig:step1}). We then unfreeze the nonlinear layers and initiate SGLD sampling (Figure \ref{fig:step2}). This procedure is designed to guide the sampler toward regions of the parameter space in which the behaviorally informed coefficients retain large magnitudes and simple hypotheses attain high log-posterior values. It also ensures that the model learns more complex representations only when they improve upon the knowledge-informed baseline. Finally, by biasing sampling toward posterior modes associated with simple hypotheses, the procedure can improve MCMC mixing by avoiding exploration of modes that are poorly supported or behaviorally implausible.

\subsection{Uncertainty Propagation from Model Parameters to Behavioral Variables}
In order for approximate Bayesian inference to be valid for behavioral variables, it must be shown how to propagate uncertainty in the model parameters propagates to uncertainty in functions of those parameters. Important behavioral variables are functions of (i) representative utilities, (ii) marginal utilities, and (iii) choice probabilities \cite{wang2020deep}. Since marginal rates of substitution are functions of marginal utilities, we focus on the posterior of the marginal-utility effect for attribute $k$ and alternative $j$ at observation $i$.\\

Let $\theta \equiv (\beta,\Theta_f,\Theta_g)$ denote all model parameters and define the marginal-utility effect as the deterministic function:
\begin{equation}
m_{ijk}(\theta)
\;\equiv\;
\frac{\partial u_{ij}}{\partial (x_{ij})_k}\!\left(\bm x_{ij}, \bm q_{ij}; \theta\right)
=
\beta_k
+ \sigma_{\text{IIA}}\,
\frac{\partial f_{\Theta_f}(\bm x_{ij}, \bm q_{ij})}{\partial [\bm x_{ij}]_k}
+ \sigma_{\text{non-IIA}}\,
\frac{\partial g_{\Theta_g}\!\left((\bm x_i, \bm q_i)_j\right)}{\partial [\bm x_{ij}]_k}.
\end{equation}
Here, $\sigma_{\text{IIA}}$ and $\sigma_{\text{non-IIA}}$ are fixed because they are treated as hyperparameters during learning. A point estimate (e.g., a local MAP within a convergence mode) is obtained by evaluating the above expression at $\hat{\theta}$.\\

A Bayesian analysis, however, requires the \emph{posterior distribution} of $m_{ijk}$ induced by the posterior over parameters. Since $m_{ijk}$ is a deterministic function of $\theta$, its posterior is the pushforward of $p(\theta\mid D)$ through $m_{ijk}(\cdot)$:
\begin{equation}
\mathbb{P}\!\left(m_{ijk}\in B\mid \bm x_{ij}, \bm q_{ij}, D\right)
=
\int \mathrm{1}\left(m_{ijk}(\theta)\in B\right) p(\theta\mid D)\, d\theta,
\end{equation}

With SGLD (and MCMC approximations in general), we obtain approximate posterior samples $\{\theta_t\}_{t=1}^T$ and propagate uncertainty by evaluating the effect at each draw,
\begin{equation}
m^{(t)}_{ijk} \;=\; m_{ijk}(\theta_t), \qquad \theta_t \sim p(\theta\mid D),
\end{equation}
which yields the empirical approximation
\begin{equation}
\mathbb{P}\!\left(m_{ijk}\in B\mid \bm x_{ij}, \bm q_{ij}, D\right)
\;\approx\;
\frac{1}{T}\sum_{t=1}^{T} \mathrm{1}\left(m_{ijk}(\theta^{(t)})\in B\right).
\end{equation}
Posterior moments can be computed by Monte Carlo integration, e.g.,
\begin{equation}
\mathbb{E}\!\left[m_{ijk}\mid \bm x_{ij}, \bm q_{ij}, D\right]
\;\approx\;
\frac{1}{T}\sum_{t=1}^{T} m^{(t)}_{ijk},
\qquad
\mathrm{Var}\!\left(m_{ijk}\mid \bm x_{ij}, \bm q_{ij}, D\right)
\;\approx\;
\frac{1}{T-1}\sum_{t=1}^{T}\left(m^{(t)}_{ijk}-\bar m_{ijk}\right)^{2},
\end{equation}
where $\bar m_{ijk} = \mathbb{E}\!\left[m_{ijk}\mid \bm x_{ij}, \bm q_{ij}, D\right]
\;$. Finally, the $(1-\alpha)$ credible interval for the marginal-utility effect is computed using posterior quantiles:
\begin{equation}
\mathrm{CI}_{1-\alpha}\!\left(m_{ijk}\right)
\;\approx\;
\Big[\, Q_{\alpha/2}\!\left(\{m^{(t)}_{ijk}\}_{t=1}^{T}\right),\;
       Q_{1-\alpha/2}\!\left(\{m^{(t)}_{ijk}\}_{t=1}^{T}\right)\,\Big].
\end{equation}
The same procedure applies to any behavioral quantity that is a deterministic function of parameters and covariates, including marginal rates of substitution (as functions of marginal utilities) and quantities derived from choice probabilities.

\subsection{Epistemic and Aleatoric Uncertainty}
In this study, we focus on uncertainty representation and leave the disentanglement of epistemic and aleatoric uncertainty outside the scope of our analysis. Recent studies have shown that sources of uncertainty in deep learning can be disentangled using multi-step learning procedures, together with architectures designed to model aleatoric uncertainty; when coupled with methods such as SGLD, these approaches could enable uncertainty disentanglement. \\

In this respect, we find the preliminary work by Yi \cite{yi2025cooperative} very promising. Although it is outside the scope of this paper, we believe that their approach could be used in conjunction with ours, and we plan to explore this synergy in future work.

\section{Monte Carlo Study for Empirical Coverage}

To estimate the performance of our model, we employ a Monte Carlo simulation study, similar to the one presented in \cite{sarrias2018individual} for Latent Class discrete choice models. Here, we assume that the latent utility function follows the equation:

\begin{equation}
  u_{ij} = \bm{x}_{ij}^{\top} \bm{\beta} + \bm{q}_i^{\top} \bm{\gamma}_j + h(\bm{x}_{i}, \bm{q}_i; \bm{\Phi}) + \epsilon_{ij}  
\end{equation} 

where $\epsilon_{ij}$ is an i.i.d. standard extreme value (EV1) random variable, $\bm{x}_{ij}$ follows an i.i.d. standard multivariate normal distribution of size five, and each vector $q_i$ consists of five dummy variables that come from i.i.d. Bernoulli distributions with $p=0.5$. Additionally, $h(\bm{x}_{ij}, \bm{q}_{i}; \bm{\Phi})$ is a non-linear function with fixed parameters $\bm{\Phi}$. The parameter vectors $\bm{\beta}, \bm{\gamma}$ are fixed, learnable parameters. \\

We simulated $D = 100$ datasets, with $J = 3$ alternatives, for $N = 1000$ and $N = 10000$ observations. For each dataset, we randomly sampled values for $\bm{x}_{ij}$, $\bm{q}_i$, and $\epsilon_{ij}$. For the data-generating process, $\bm{\gamma}_j = 0$ for alternatives $j = 1$ and $j = 2$. Regarding $h(\bm{x}_{ij}, \bm{q}_i; \bm{\Phi})$, we included a third-degree polynomial and a hyperbolic tangent nonlinearity that depends on a subset of $\bm x_{ij}$ to enable nonlinear--as well as some strictly linear--dependencies in the latent utilities. We selected the parameters $\bm{\Phi}$ to ensure reasonably good performance (around 70\% weighted out-of-sample accuracy) of the knowledge-informed part of the model on the simulated datasets.\footnote{We did this to ensure a more than fair comparison of our model against models that ignore the nonlinear component of the latent utilities.}\\

Since, in discrete choice, we are primarily interested in inference regarding the alternative-specific attributes $\bm{x}$, we estimate the derivatives of utility $u_i$ with respect to these inputs for each case to obtain estimates for the marginal rates of substitution. We then compare them with the true marginal rates of substitution, computed using the known function parameters and their derivatives. \\

To assess the reliability of these estimates, we compute the empirical coverage, defined as the proportion of simulated datasets for which the average marginal rates of substitution (MRS) fall within the corresponding estimated credible intervals. We also show credible bands for the latent utilities to illustrate how well the models approximate the latent function at various values of the alternative attributes. Additionally, we compute the out-of-sample accuracy of the models’ predictions based on the average of the SGLD samples for the latent utilities of each alternative. \\

\section{Revealed and Stated Preference Case Studies}

In addition to the simulation study, we consider two case studies: one with revealed preference data and another with stated preference data. We use the revealed preference dataset constructed in \cite{VILLARRAGA2025103132} for multinomial mode choice in NYC. For the stated preference data, we use the popular and publicly available\footnote{Available in Apollo in R as the \texttt{apollo\_swissRouteChoiceData}.} dataset from a binary train choice experiment in Switzerland \cite{vrtic2002impact}. \\

\subsection{Mode Choice in New York City}
 We select home-based work (HBW) trips completed in transit, private car, or non-motorized modes (i.e. walking and bicycling) without limiting the trip length. Trip cost for the non-motorized alternative is set to zero and the travel time is computed using the Google API for the cases where it is not revealed. We discarded observations with missing data, so after the data selection and cleaning process, 3,277 trips are left for our analysis. The description for the variables considered in the multinomial problem is presented in Table \ref{tab:summary_multi}.
\begin{table}[h!]
	\centering 
	\caption{Summary of the variables considered for the multinomial mode choice problem.}
	\begin{tabular}{p{5cm}p{8cm}S[table-format=3.2]}
		\hline 
		\centering Variable & \centering  Description & \multicolumn{1}{c}{Mean}  \\
		\hline
		\centering Trip cost transit &  Transit cost (USD). &  2.35 \\
          \centering Trip cost car &  Car cost (USD). &  4.64 \\
          \centering Trip cost non-motorized &  Non-motorized cost (USD). & 0.00  \\
		\centering Trip time transit &  Transit travel time (Minutes). & 50.38 \\
  \centering Trip time car &  Car travel time (Minutes). & 18.32 \\
  \centering Trip time non-motorized &  Non-motorized travel time (Minutes). & 93.85 \\
		\centering Vehicle availability &  Indicator variable for car availability in the household. It takes the value of 1 if no cars are available in the household. & 0.36 \\
		\centering High income &  Indicator variable for high income level ($>100$k USD per year). & 0.32 \\
		\centering Manhattan &  Indicator variable for destinations in Manhattan. &  0.46 \\
		\centering Gender &  Indicator variable for the male gender. & 0.47  \\
		\centering \textbf{Car mode share} &  Proportion of trips completed by car & 0.37 \\
      \centering \textbf{Transit mode share} &  Proportion of trips completed by transit & 0.51 \\
      \centering \textbf{Non-motorized mode share} &  Proportion of trips completed by non-motorized means of transportation & 0.12 \\
		\hline
	\end{tabular}
	\label{tab:summary_multi}
\end{table}

\subsection{Swiss Train Data}

This dataset contains information on two train route alternatives. It includes alternative-specific attributes such as travel time, trip cost, headway time, and the number of interchanges. Additionally, socio-demographic variables, including household income and trip purpose, are also included. A summary and description of all variables are presented in Table \ref{tab:summary_swiss_train}. \\

\begin{table}[h!]
	\centering 
	\caption{Summary of the variables considered for the Swiss train dataset.}
	\begin{tabular}{p{5.5cm}p{7cm}c}
		\hline 
		\centering \textbf{Variable} & \centering \textbf{Description} & \multicolumn{1}{c}{\textbf{Mean}}  \\
		\hline
		\centering Choice & Choice indicator, 0 for alternative 1, and 1 for alternative 2. & 0.50 \\
		\centering Travel time (alt. 1) & Travel time (in minutes) for alternative 1. & 52.59 \\
        \centering Travel cost (alt. 1) & Travel cost (in CHF) for alternative 1. & 19.67 \\
        \centering Headway time (alt. 1) & Headway time (in minutes) for alternative 1. & 32.48 \\
        \centering Number of interchanges (alt. 1) & Number of interchanges for alternative 1. & 0.94 \\
		\centering Travel time (alt. 2) & Travel time (in minutes) for alternative 2. & 52.47 \\
        \centering Travel cost (alt. 2) & Travel cost (in CHF) for alternative 2. & 19.69 \\
        \centering Headway time (alt. 2) & Headway time (in minutes) for alternative 2. & 32.38 \\
        \centering Number of interchanges (alt. 2) & Number of interchanges for alternative 2. & 0.94 \\
		\centering Household income & Household income (in CHF per annum). & 76508 \\
		\centering Car availability & Indicator variable for car availability. It takes the value of 1 if the respondent has a car available, 0 otherwise. & 0.38 \\
		\centering Commute & Indicator variable for commuting trips (1 if commuting, 0 otherwise). & 0.29 \\
		\centering Shopping & Indicator variable for shopping trips (1 if shopping, 0 otherwise). & 0.08 \\
		\centering Business & Indicator variable for business trips (1 if business, 0 otherwise). & 0.09 \\
		\centering Leisure & Indicator variable for leisure trips (1 if leisure, 0 otherwise). & 0.54 \\
		\hline
	\end{tabular}
	\label{tab:summary_swiss_train}
\end{table}

\section{Results}

In this section, we present the results for each dataset, as well as the definition of our model for each case. We begin with the simulation study and then proceed to the results for the revealed and stated preference data.\\

\subsection{Simulation Study}

\begin{figure}[h!]
    \centering
    \begin{subfigure}[t]{0.45\textwidth}
        \centering
        \includegraphics[width=\linewidth]{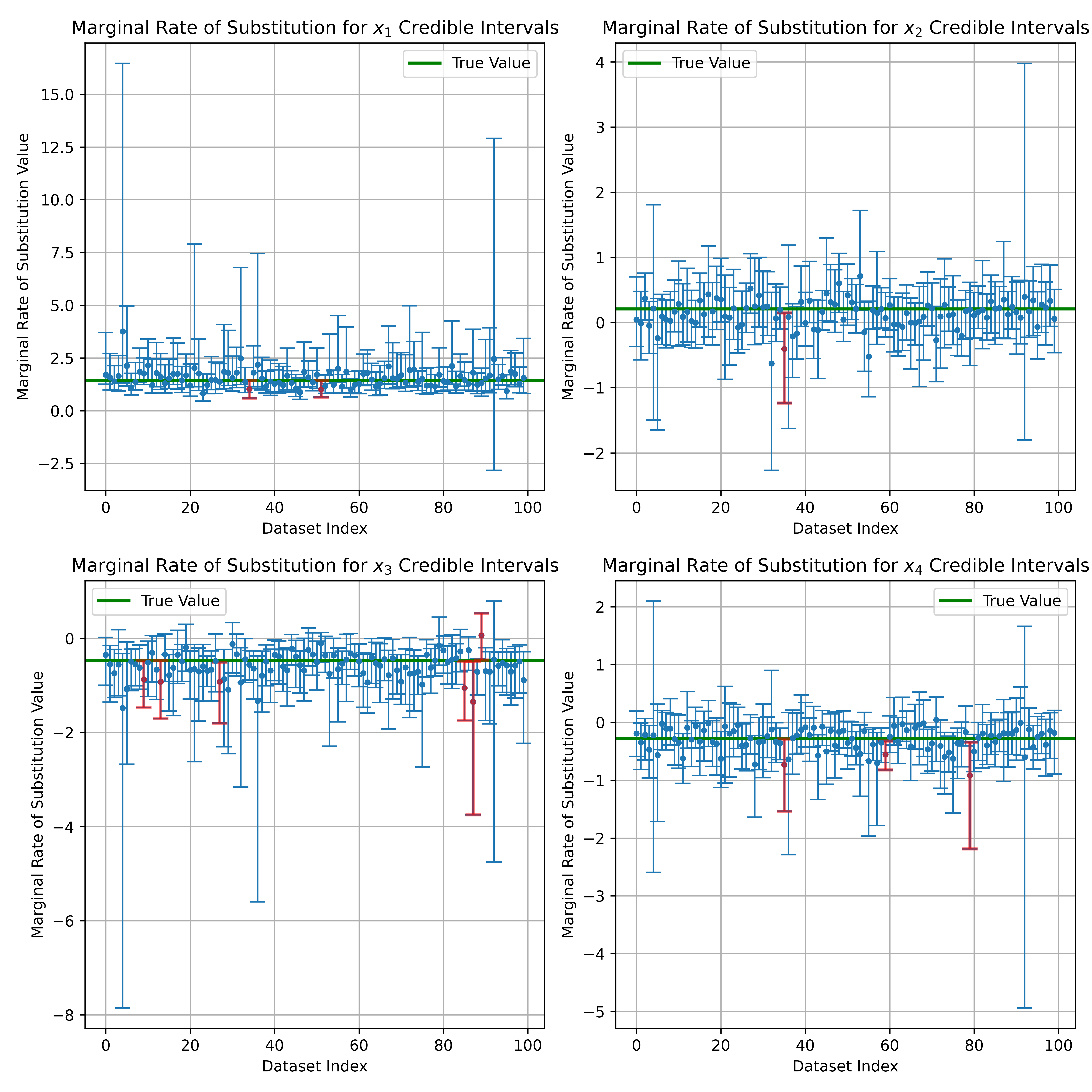}
        \caption{Dataset size: 1000}
        \label{fig:mrs_ci_iia_skip_model_per_dataset_1000}
    \end{subfigure}
    \hfill
    \begin{subfigure}[t]{0.45\textwidth}
        \centering
        \includegraphics[width=\linewidth]{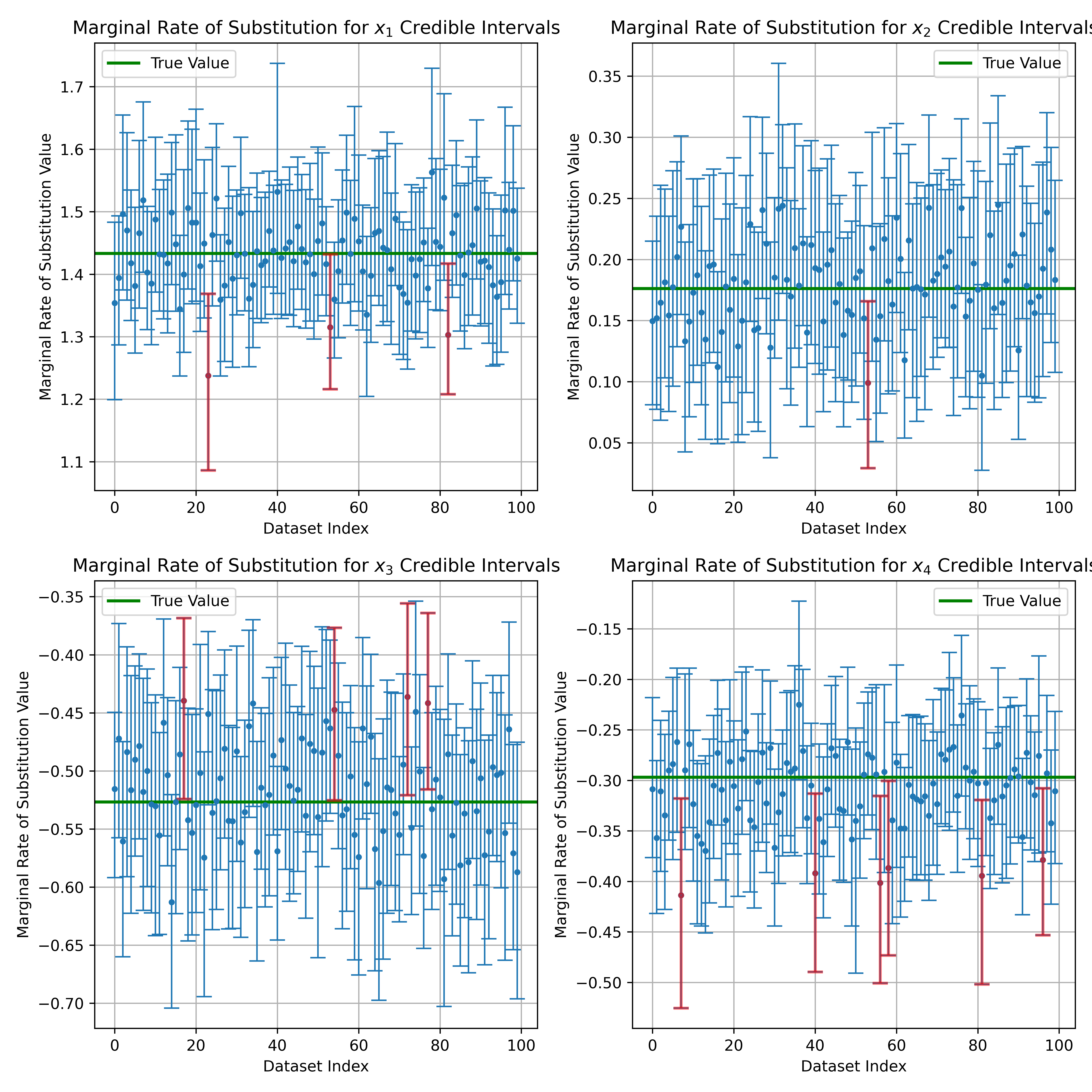}
        \caption{Dataset size: 10000}
        \label{fig:mrs_ci_iia_skip_model_per_dataset_10000}
    \end{subfigure}
    \caption{Credible intervals for marginal rates of substitution across individual simulated datasets using our proposed model. Red intervals indicate cases where the true value is not captured.}
    \label{fig:mrs_ci_iia_skip_model_per_dataset_all}
\end{figure}

\begin{figure}[h!]
    \centering
    \begin{subfigure}[t]{0.45\textwidth}
        \centering
        \includegraphics[width=\linewidth]{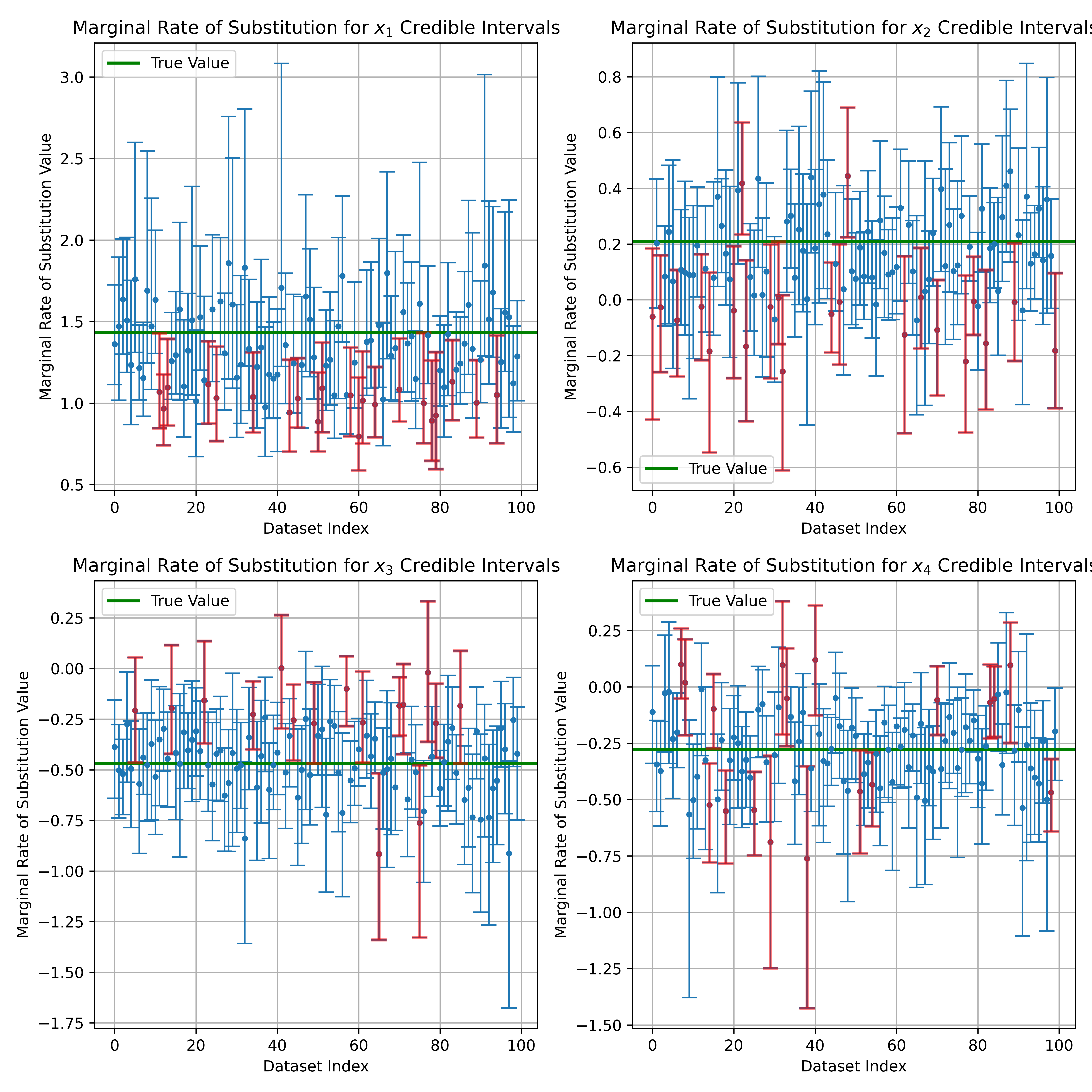}
        \caption{Dataset size: 1000}
        \label{fig:mrs_ci_iia_fully_nn_per_dataset_1000}
    \end{subfigure}
    \hfill
    \begin{subfigure}[t]{0.45\textwidth}
        \centering
        \includegraphics[width=\linewidth]{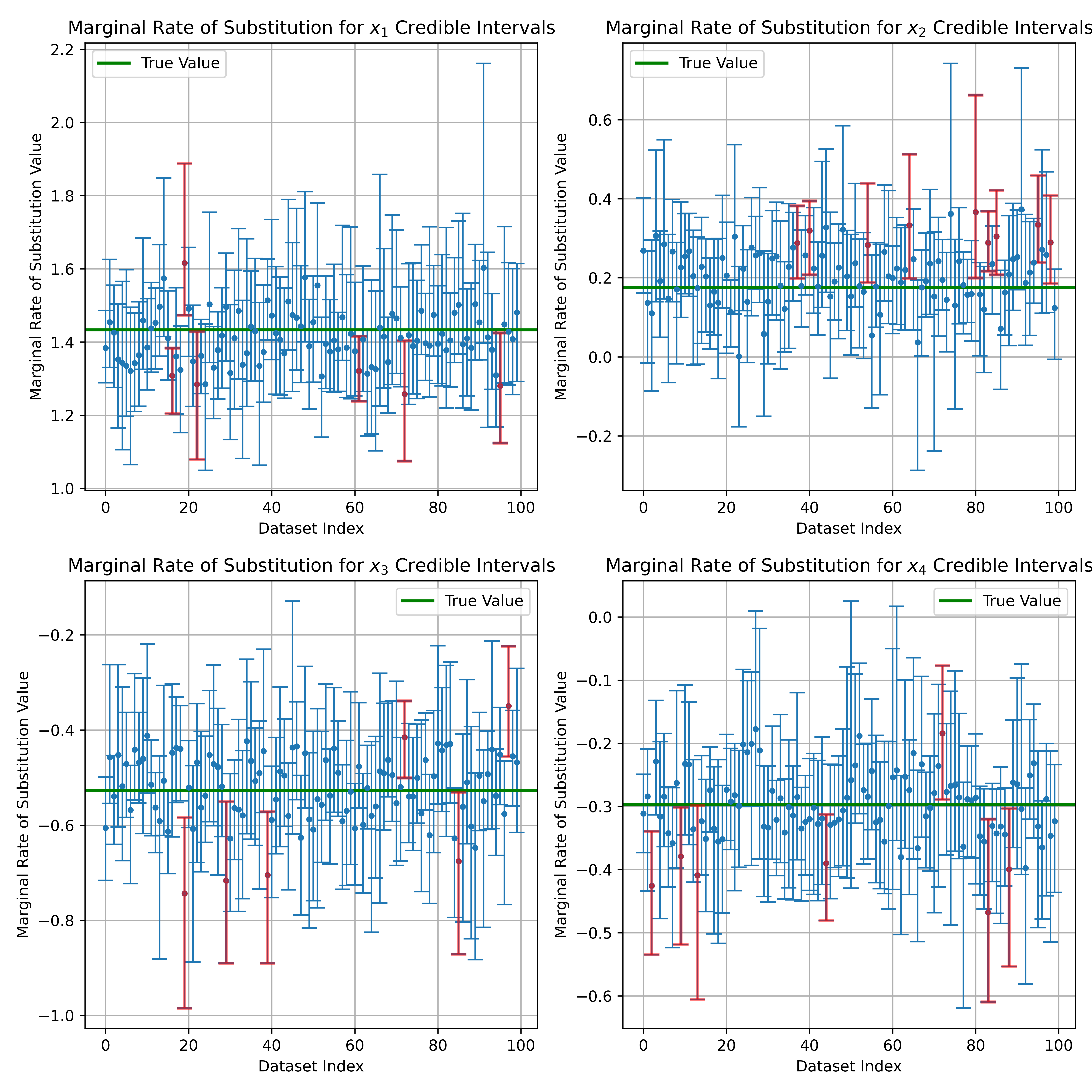}
        \caption{Dataset size: 10000}
        \label{fig:mrs_ci_iia_fully_nn_per_dataset_10000}
    \end{subfigure}
    \caption{Credible intervals for marginal rates of substitution across individual simulated datasets using the fully connected neural network. Red intervals indicate cases where the true value is not captured.}
    \label{fig:mrs_ci_iia_fully_nn_per_dataset_all}
\end{figure}

\begin{figure}[h!]
    \centering
    \begin{subfigure}[t]{0.45\textwidth}
        \centering
        \includegraphics[width=\linewidth]{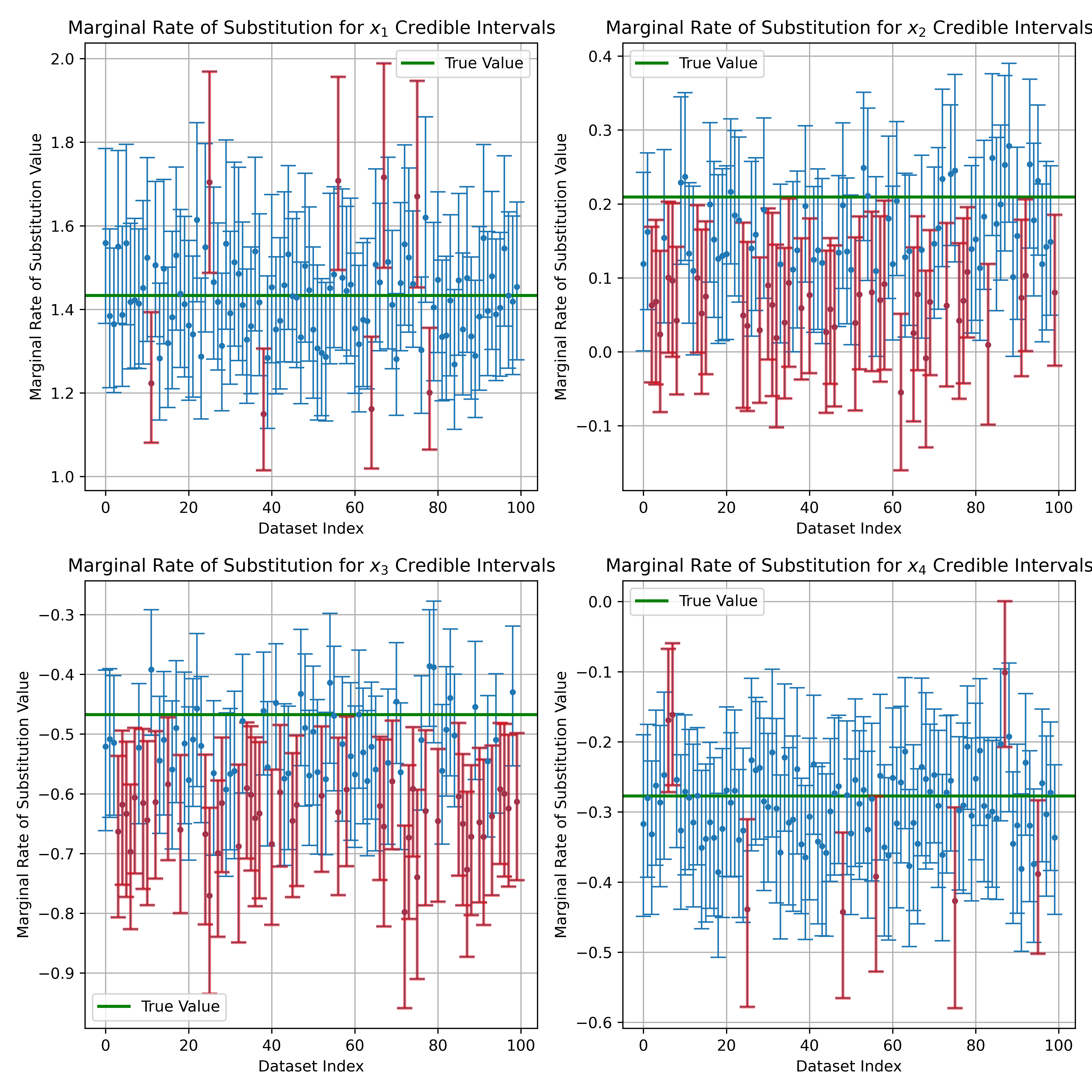}
        \caption{Dataset size: 1000}
        \label{fig:mrs_ci_iia_linear_per_dataset_1000}
    \end{subfigure}
    \hfill
    \begin{subfigure}[t]{0.45\textwidth}
        \centering
        \includegraphics[width=\linewidth]{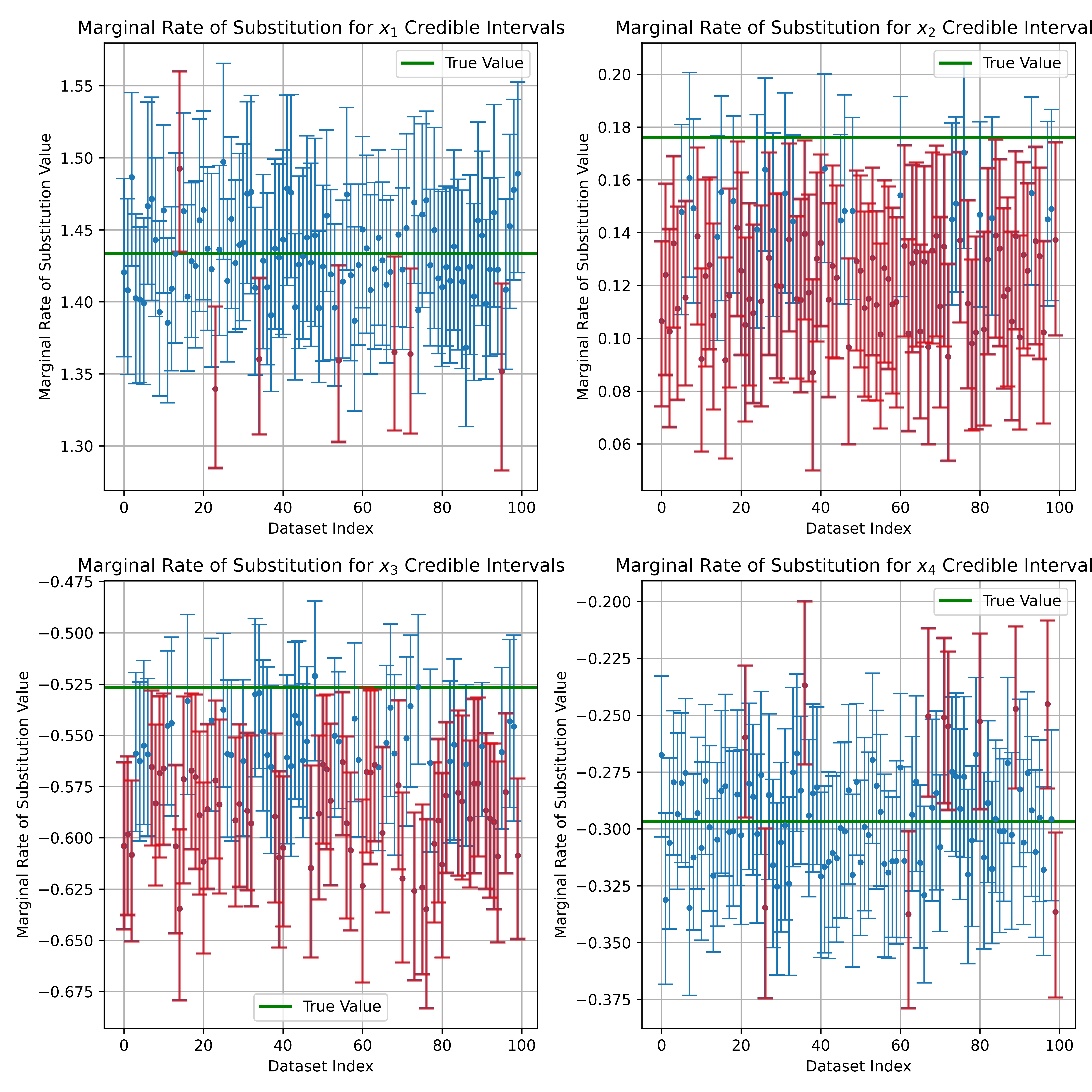}
        \caption{Dataset size: 10000}
        \label{fig:mrs_ci_iia_linear_per_dataset_10000}
    \end{subfigure}
    \caption{Credible intervals for marginal rates of substitution across individual simulated datasets using the conditional logit model. Red intervals indicate cases where the true value is not captured.}
    \label{fig:mrs_ci_iia_linear_per_dataset_all}
\end{figure}

For the simulation study, we computed marginal rates of substitution (MRS) with respect to the first alternative attribute. Since the study includes five alternative attributes, we are able to compute four MRS values. In Figure~\ref{fig:mrs_ci_iia_skip_model_per_dataset_all}, we present the 95\% credible intervals for these four MRS across all simulated datasets for $N = 1000$ and $N = 10000$ observations using our model. In Figure~\ref{fig:mrs_ci_iia_fully_nn_per_dataset_all}, we show the same results for a fully connected neural network, and in Figure~\ref{fig:mrs_ci_iia_linear_per_dataset_all}, we present the results for a model that ignores the nonlinear components. \footnote{For all neural networks—both in our model and in the fully connected model—we used 512 hidden units, two hidden layers, and ReLU activations. The $\ell_2$ regularization terms were set to 0.0001.} \\

As shown in these figures, our model achieves empirical coverage consistent with the 95\% credibility level, with very few datasets in which the credible intervals fail to capture the true average MRS. For the fully connected neural network, we observe that for $N = 1000$ (Figure~\ref{fig:mrs_ci_iia_fully_nn_per_dataset_1000}), there are many instances where the credible intervals fail to capture the true MRS, regardless of which specific MRS is being analyzed. For $N = 10000$ (Figure~\ref{fig:mrs_ci_iia_fully_nn_per_dataset_10000}), the fully connected neural network attains higher empirical coverage. In contrast, for the model that ignores the nonlinear component, the credible intervals frequently fail for the second and third MRS—those associated with variables that exhibit highly nonlinear relationships with the latent utilities, as we will discuss later in this section. These results are formally summarized in Table~\ref{tab:empirical_coverage}, where we report the empirical coverage for each model, dataset size, and MRS.\\

The empirical coverages for the MRS, illustrated in Figures~\ref{fig:mrs_ci_iia_skip_model_per_dataset_all} to~\ref{fig:mrs_ci_iia_linear_per_dataset_all} and summarized in Table~\ref{tab:empirical_coverage}, show that:  
i) our model achieves empirical coverage consistent with a 95\% credibility level, regardless of dataset size;  
ii) the fully connected neural network fails to achieve adequate coverage for the smaller datasets but attains coverage close to 95\% for the larger datasets; and iii) the model that ignores nonlinearities (the conditional logit model) achieves coverage close to 95\% only for MRS associated with alternative attributes that have linear or near-linear relationships with the latent utilities, as we will show next\footnote{Alternative attributes $x_1$ and $x_2$ have a linear relationship with the latent utilities, so the first MRS should be accurately estimated by a linear model.}.\\ 

\begin{table}[h!]
    \centering
    \caption{Empirical coverage of average marginal rates of substitution across 100 simulations. Results obtained using SGLD with 5,000 epochs.}
    \begin{tabular}{cc|cccc|c}
    \hline
    Model & Dataset Size & MRS 1 & MRS 2 & MRS 3 & MRS 4 & Average \\
    \hline
    Our model & 1000 & 98\% & 99\% & 94\% & 97\% & 97.0\% \\
    Our model & 10000 & 97\% & 99\% & 96\% & 94\% & 96.5\% \\
    Fully connected NN & 1000 & 79\% & 78\% & 84\% & 82\% & 80.8\% \\
    Fully connected NN & 10000 & 94\% & 91\% & 94\% & 93\% & 93.0\% \\
    Conditional Logit Model & 1000 & 92\% & 60\% & 54\% & 92\% & 74.5\% \\
    Conditional Logit Model & 10000 & 93\% & 24\% & 43\% & 89\% & 62.3\% \\
    \hline
    \end{tabular}
    \label{tab:empirical_coverage}
\end{table}

\begin{figure}[h!]
    \centering
    \begin{subfigure}[t]{0.45\textwidth}
        \centering
        \includegraphics[width=\linewidth]{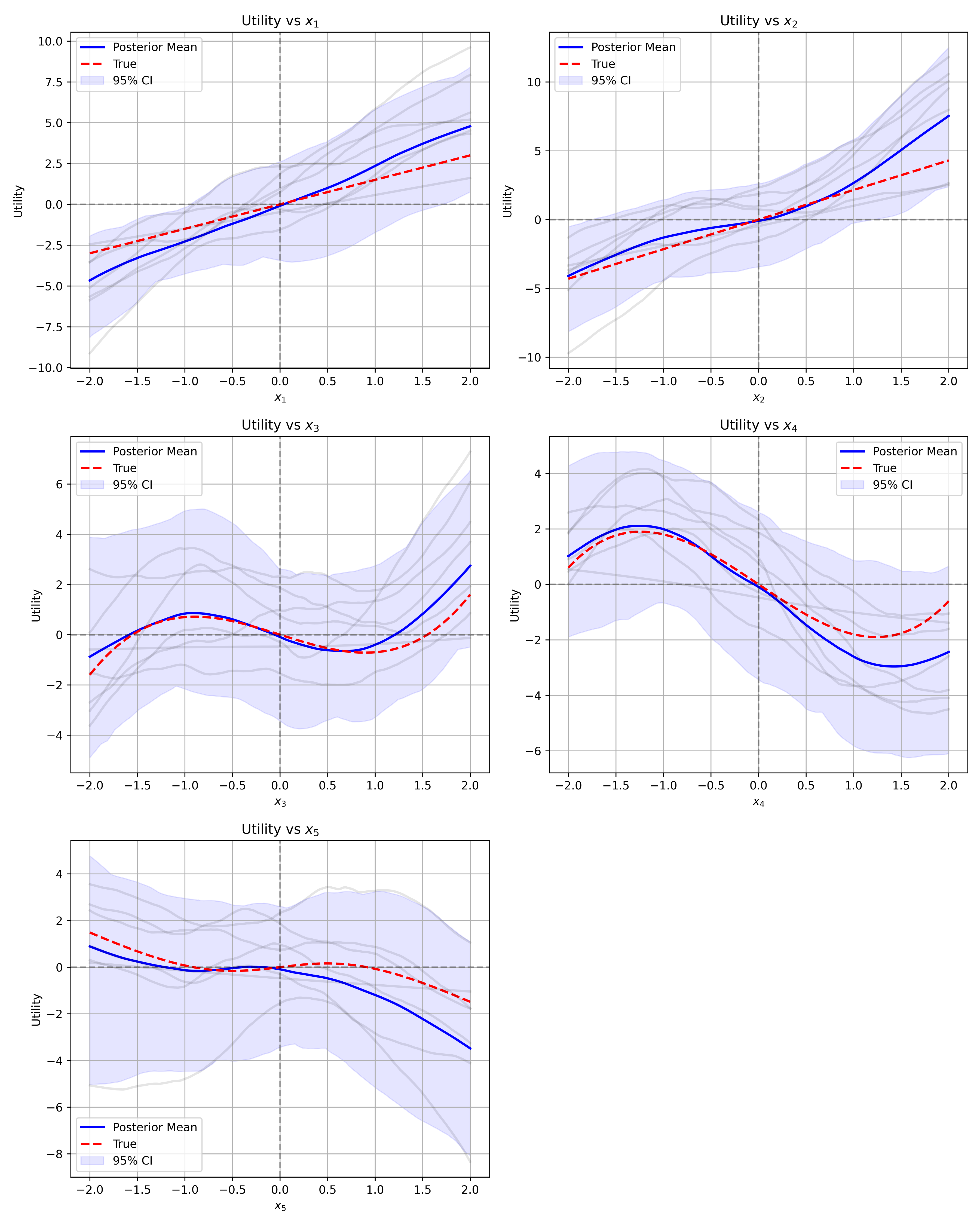}
        \caption{Dataset size: 1000}
        \label{fig:utility_predictions_skip_model_1000_high_nonlinearity}
    \end{subfigure}
    \hfill
    \begin{subfigure}[t]{0.45\textwidth}
        \centering
        \includegraphics[width=\linewidth]{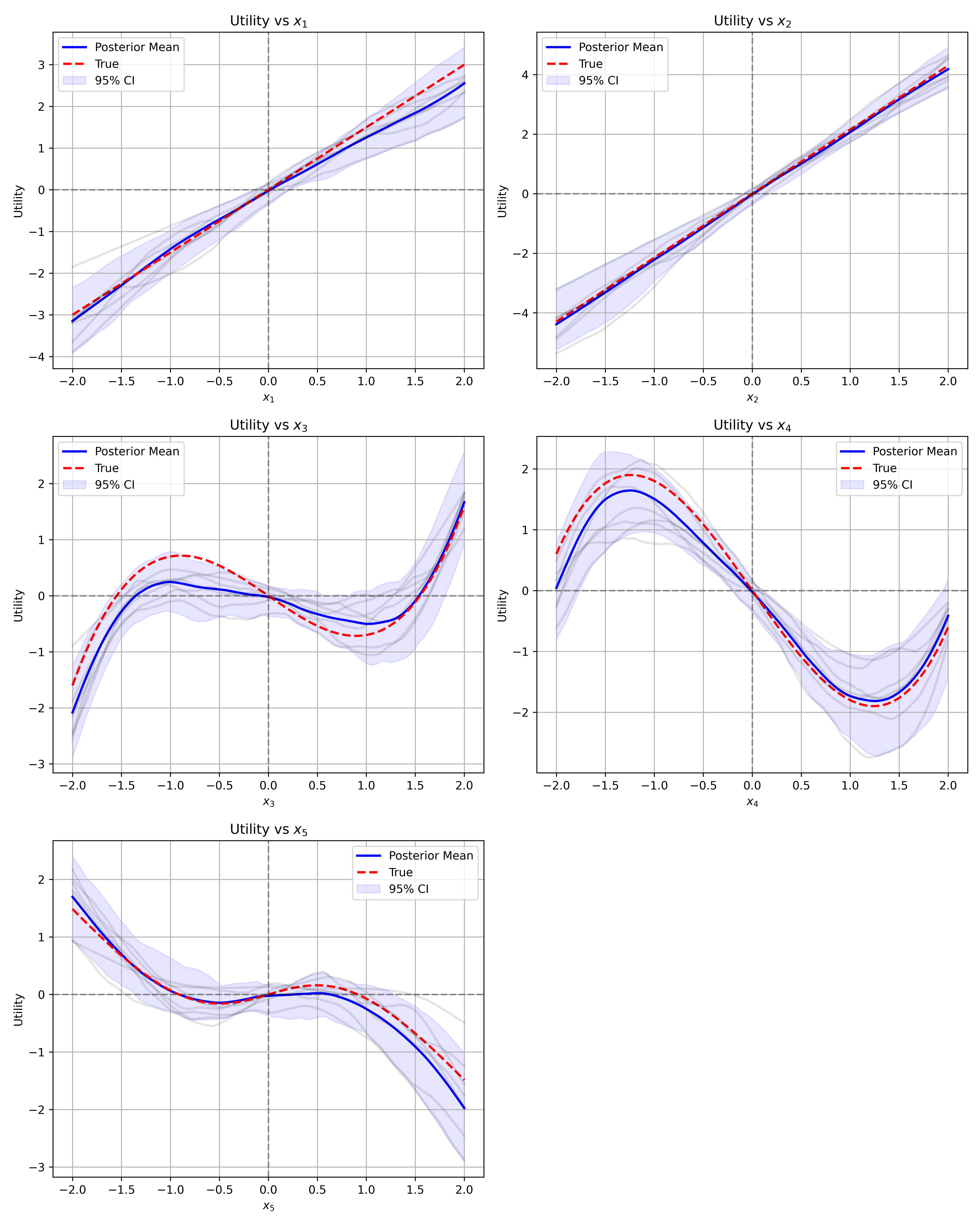}
        \caption{Dataset size: 10000}
        \label{fig:utility_predictions_skip_model_10000}
    \end{subfigure}
    \caption{Representative utility predictions as a function of the alternative attributes for our proposed model. 95\% Credible intervals are depicted in shaded blue. True representative utilities are presented in dashed red.}
    \label{fig:utility_predictions_skip_model_all}
\end{figure}

\begin{figure}[h!]
    \centering
    \begin{subfigure}[t]{0.45\textwidth}
        \centering
        \includegraphics[width=\linewidth]{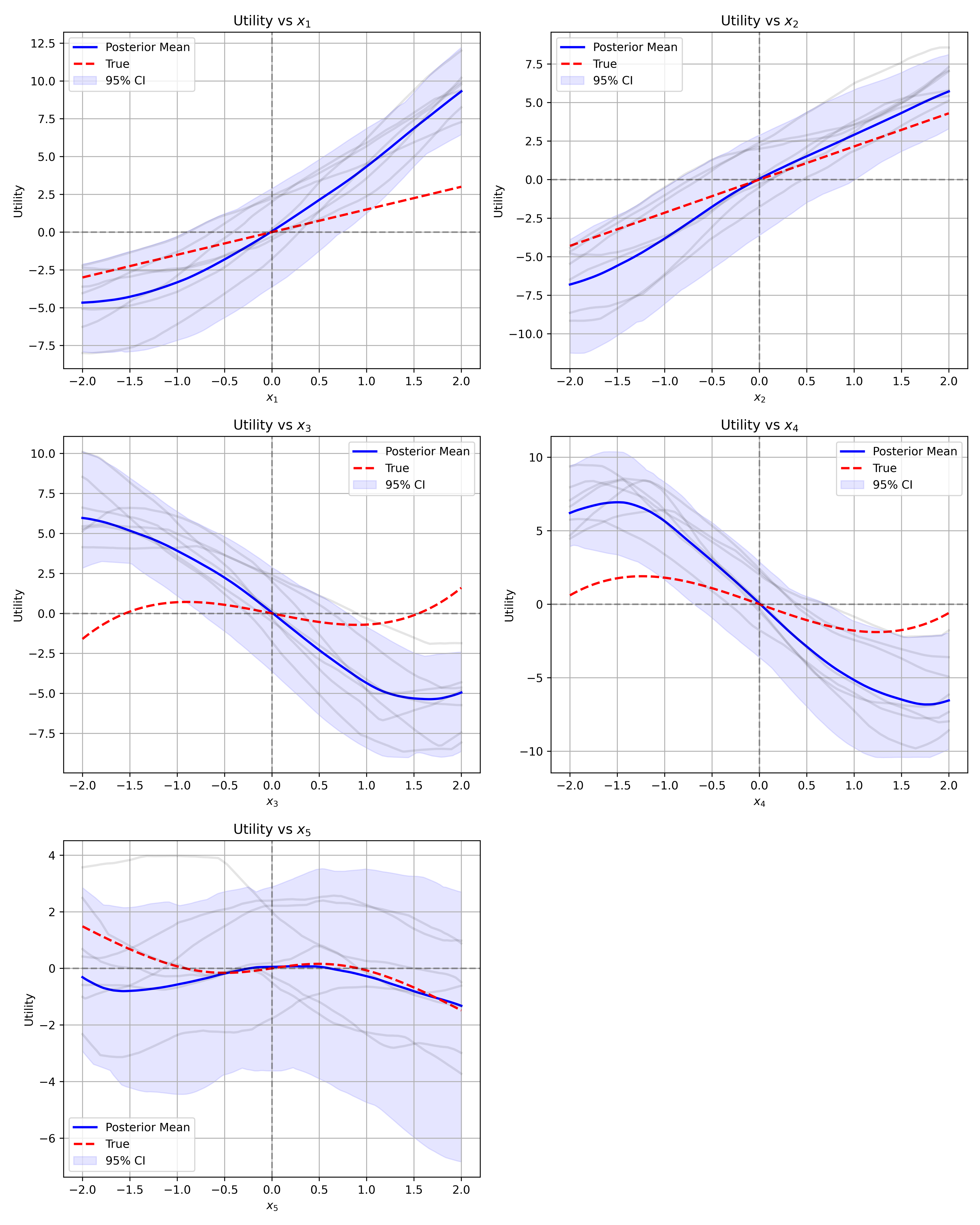}
        \caption{Dataset size: 1000}
        \label{fig:utility_predictions_fully_nn_1000_high_nonlinearity}
    \end{subfigure}
    \hfill
    \begin{subfigure}[t]{0.45\textwidth}
        \centering
        \includegraphics[width=\linewidth]{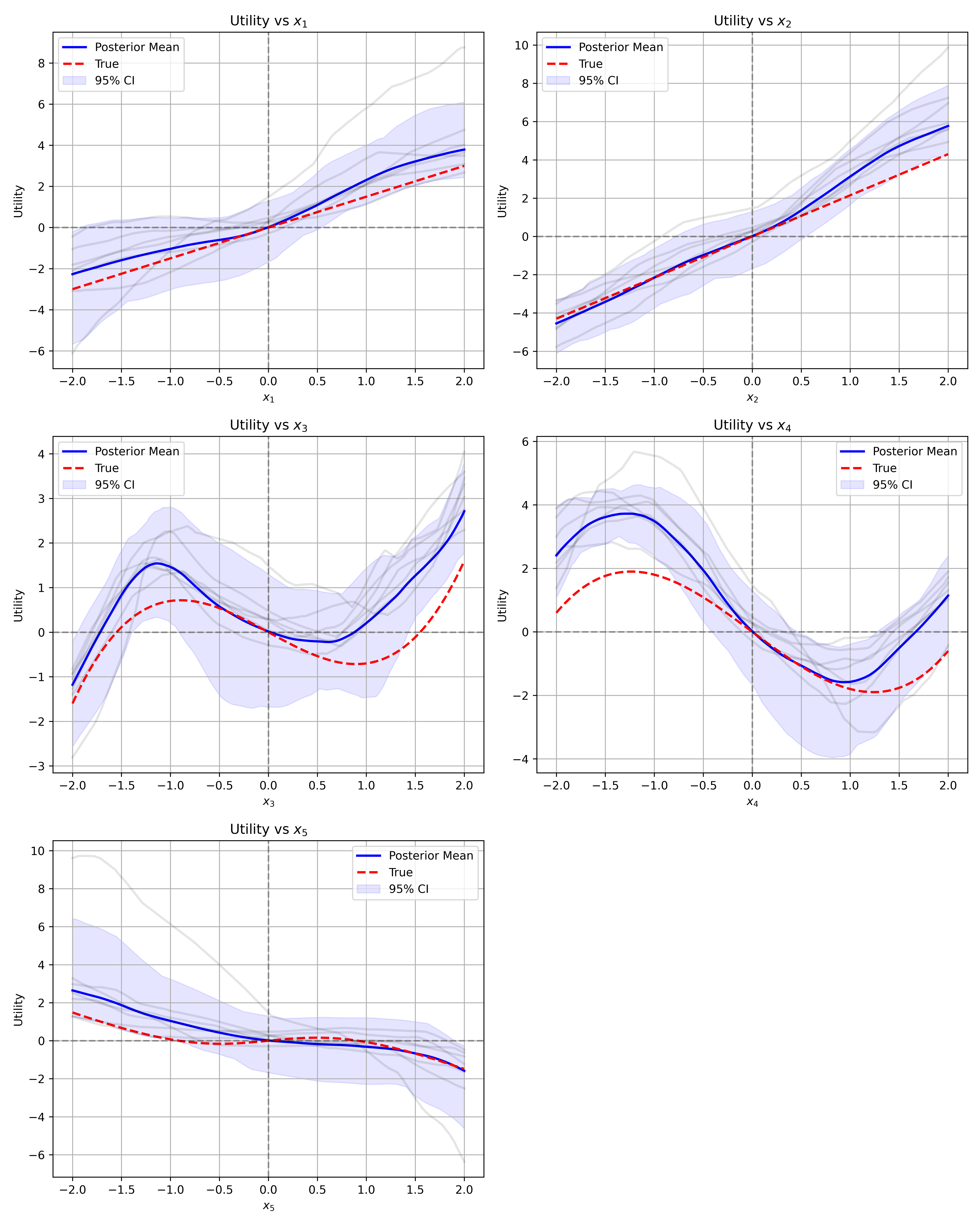}
        \caption{Dataset size: 10000}
        \label{fig:utility_predictions_fully_nn_10000}
    \end{subfigure}
    \caption{Representative utility predictions as a function of the alternative attributes for the fully connected neural network. 95\% Credible intervals are depicted in shaded blue. True representative utilities are presented in dashed red.}
    \label{fig:utility_predictions_fully_nn_all}
\end{figure}

\begin{figure}[h!]
    \centering
    \begin{subfigure}[t]{0.45\textwidth}
        \centering
        \includegraphics[width=\linewidth]{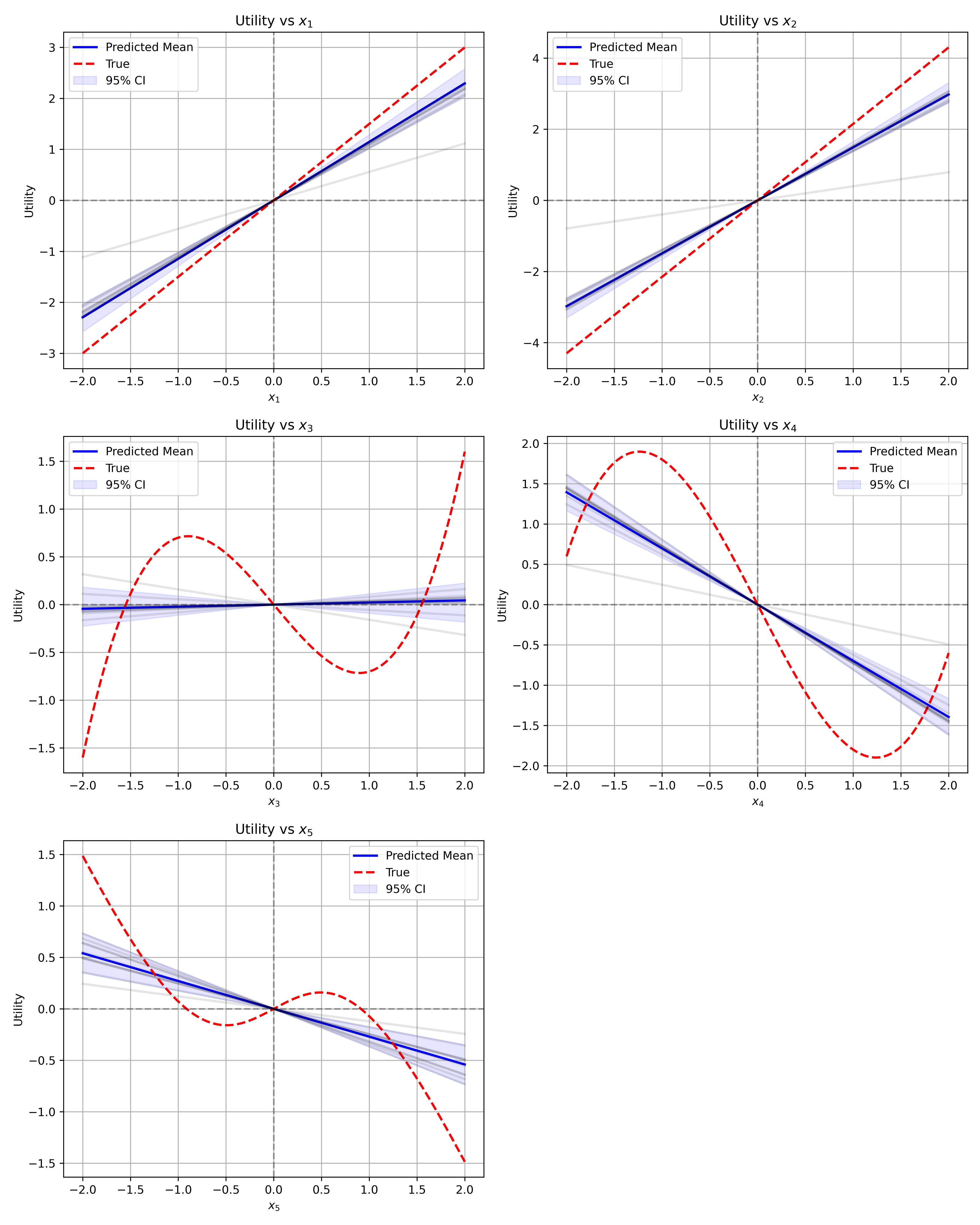}
        \caption{Dataset size: 1000}
        \label{fig:utility_predictions_linear_1000_high_nonlinearity}
    \end{subfigure}
    \hfill
    \begin{subfigure}[t]{0.45\textwidth}
        \centering
        \includegraphics[width=\linewidth]{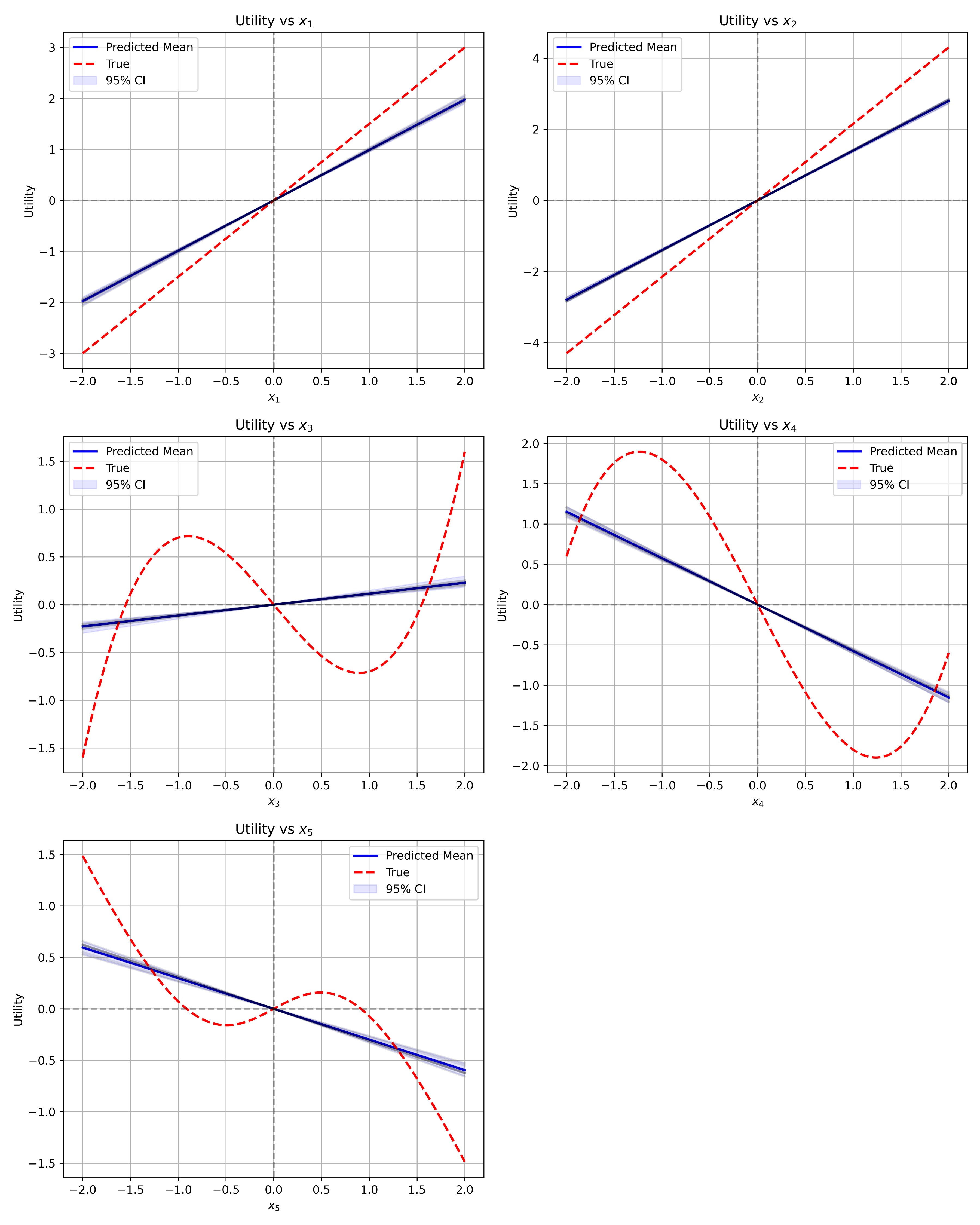}
        \caption{Dataset size: 10000}
        \label{fig:utility_predictions_linear_10000}
    \end{subfigure}
    \caption{Representative utility predictions as a function of the alternative attributes for the conditional logit model. 95\% Credible intervals are depicted in shaded blue. True representative utilities are presented in dashed red.}
    \label{fig:utility_predictions_linear_all}
\end{figure}

To better assess how well our model approximates the latent utilities—and to contrast it with a fully connected neural network and a conditional logit model—we plot the 95\% credible bands for the latent utilities as a function of each alternative attribute, for every model and dataset size. In Figure~\ref{fig:utility_predictions_skip_model_all}, we present these results for our model; in Figure~\ref{fig:utility_predictions_fully_nn_all}, we show the results for the fully connected neural network; and in Figure~\ref{fig:utility_predictions_linear_all}, we show those associated with the conditional logit model.\\

For our model, the credible bands (Figure~\ref{fig:utility_predictions_skip_model_all}) successfully cover the true utilities for all alternative attributes and dataset sizes. Also, we observe that the posterior mean approximates very well the true latent function shape. Additionally, we observe that the credible bands become narrower as the dataset size increases, which is a good indication that the SGLD implementation reliably represents epistemic uncertainty.\\

For the fully connected neural network, on the other hand, the posterior credible bands for the latent utilities (Figure~\ref{fig:utility_predictions_fully_nn_all}) do not cover the true latent function. However, we observe that the posterior mean is able to capture sign changes in the first-order derivative of the latent function. Moreover, when the dataset size increases (Figure~\ref{fig:utility_predictions_fully_nn_10000}), these approximations improve significantly, as clearly illustrated for the latent utilities with respect to $x_3$ and $x_4$.\\

The posterior credible bands for the linear model (Figure~\ref{fig:utility_predictions_linear_all}) do not cover the true latent utilities, and the posterior mean does not approximate them well—even for attributes with a linear relationship ($x_1$ and $x_2$). More concerning, however, is that we observe very narrow credible intervals, indicating that models which ignore the nonlinear components of the latent utilities, when present, can be overconfident in their predictions and fail to adequately capture and represent uncertainty.\\ 

Finally, regarding the comparative predictive performance of our model against the fully connected neural network and the conditional logit model, we found that: i) our model achieves predictive performance equivalent to—or better than—that of the other two models across all dataset sizes; and ii) our model is capable of effectively leveraging additional data to improve predictive accuracy, as evidenced by an 8-percentage-point increase in weighted out-of-sample accuracy from $N = 1000$ to $N = 10000$. The results for weighted out-of-sample accuracy are summarized in Table~\ref{tab:accuracy_sim}.\\

\begin{table}[h!]
    \centering
    \caption{Out-of-sample accuracy for the simulation study. Predictions are computed using samples from SGLD. Results obtained using SGLD with 5,000 epochs.}
    \begin{tabular}{ccc}
    \hline
    Model & Dataset Size & Accuracy (\%)\\
    \hline
    Our model & 1000 & 70.97 \\
    Our model & 10000 & 79.00 \\
    Fully connected NN & 1000 & 69.99 \\
    Fully connected NN & 10000 & 78.02 \\
    Conditional Logit Model & 1000 & 72.52 \\
    Conditional Logit Model & 10000 & 72.91 \\
    \hline
    \end{tabular}
    \label{tab:accuracy_sim}
\end{table}
\subsection{Mode Choice in New York City}

For the revealed preference case study using New York City multinomial mode choice data, we estimated values of travel time savings (VOTT) for train and car, as well as the out-of-sample accuracy of our model and a fully connected neural network. Standard discrete choice models, graph neural networks, and hierarchical Gaussian process models were previously estimated in \cite{VILLARRAGA2025103132} and \cite{2025arXiv250309786V}.\\

\begin{figure}[h!]
    \centering
    \begin{subfigure}[t]{0.3\textwidth}
        \centering
        \includegraphics[width=\linewidth]{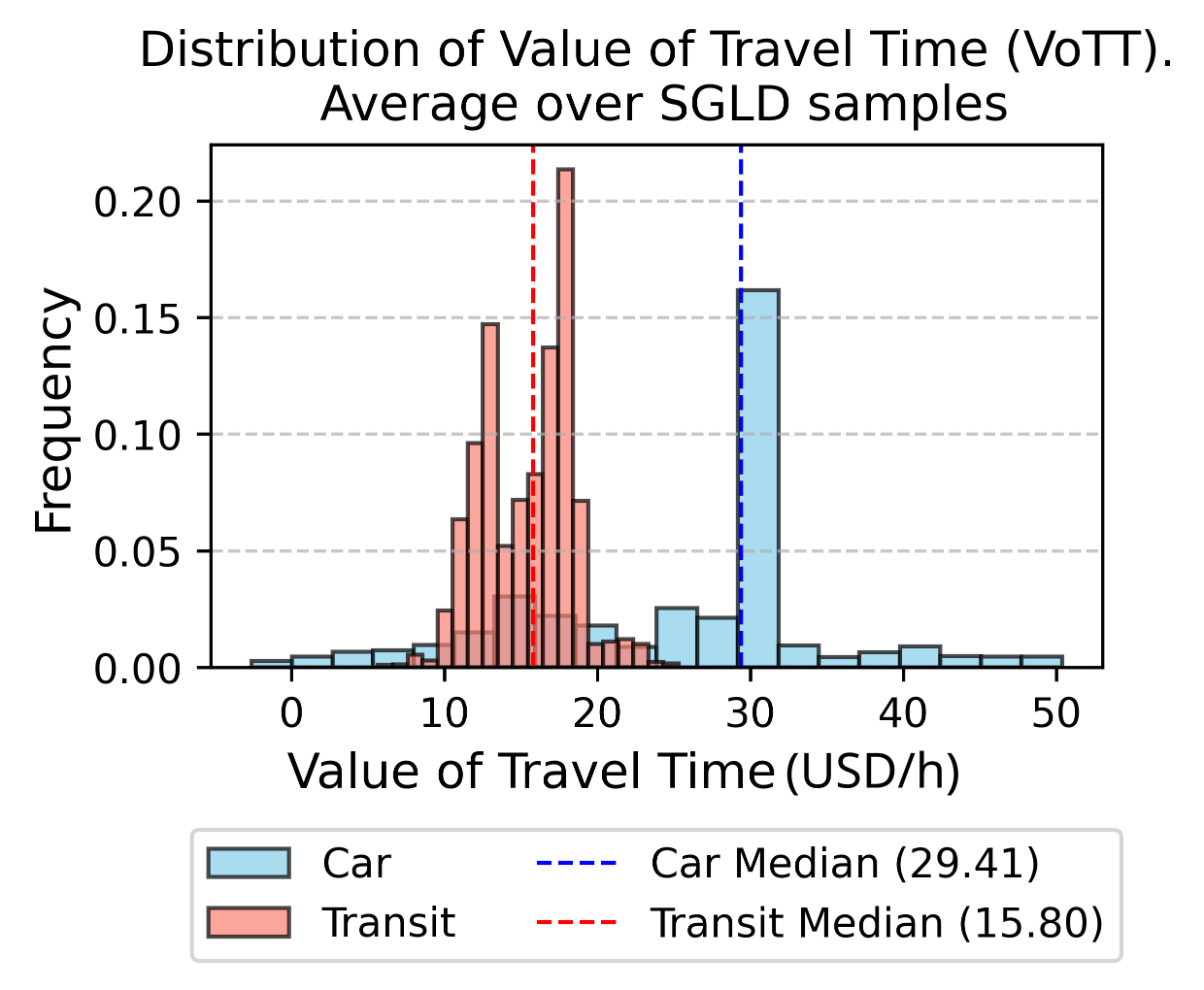}
        \caption{High $\ell_2$ penalty applied to non-linear scale parameters.}
        \label{fig:vott_skip_nn_mode_mode_high_l2}
    \end{subfigure}
    \hfill
    \begin{subfigure}[t]{0.3\textwidth}
        \centering
        \includegraphics[width=\linewidth]{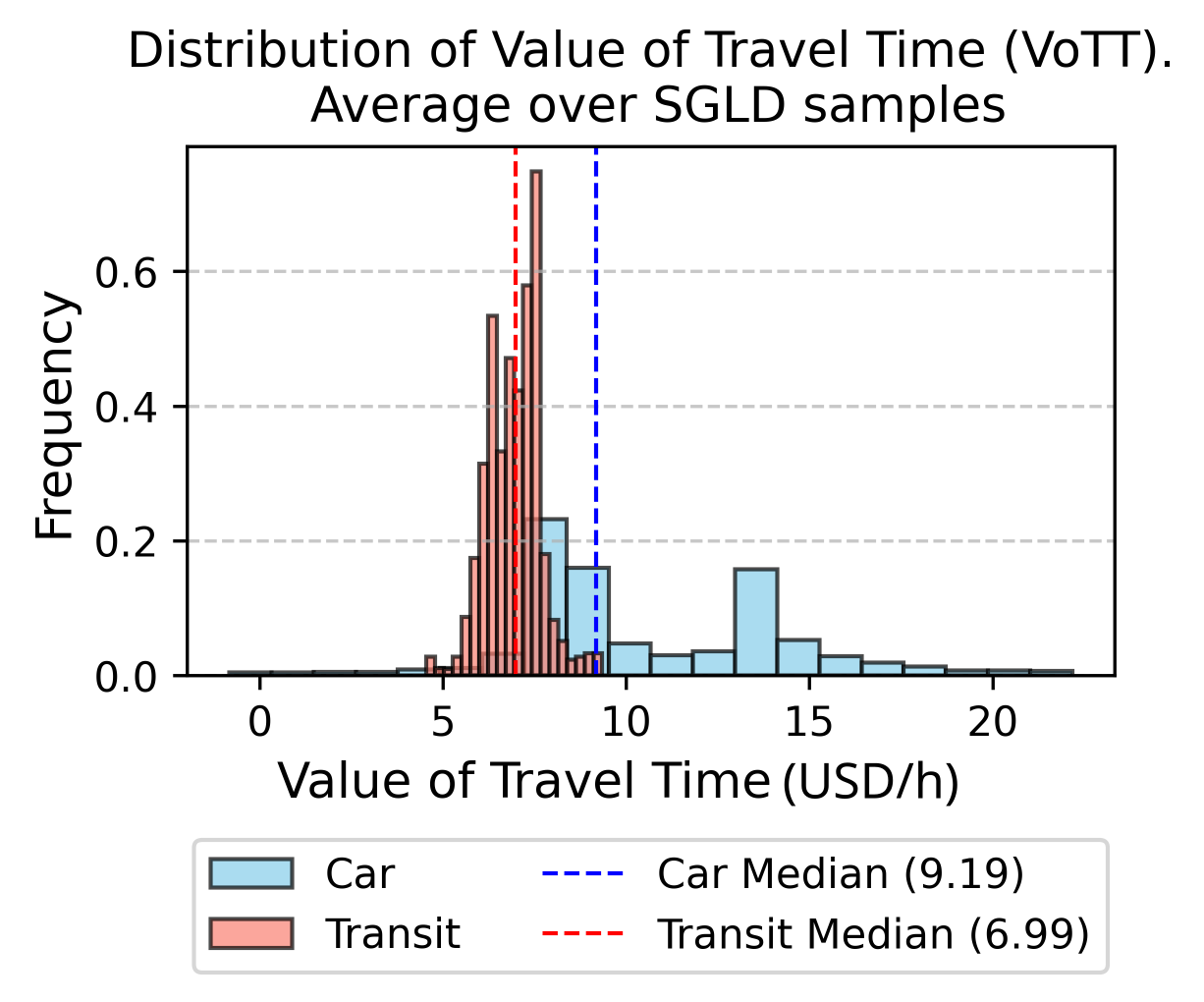}
        \caption{High $\ell_2$ penalty applied to non-linear non-IIA scale parameter. No penalty for non-linear IIA scale parameter.}
        \label{fig:vott_skip_nn_mode_mode_high_l2_non_iia}
    \end{subfigure}
    \hfill
    \begin{subfigure}[t]{0.3\textwidth}
        \centering
        \includegraphics[width=\linewidth]{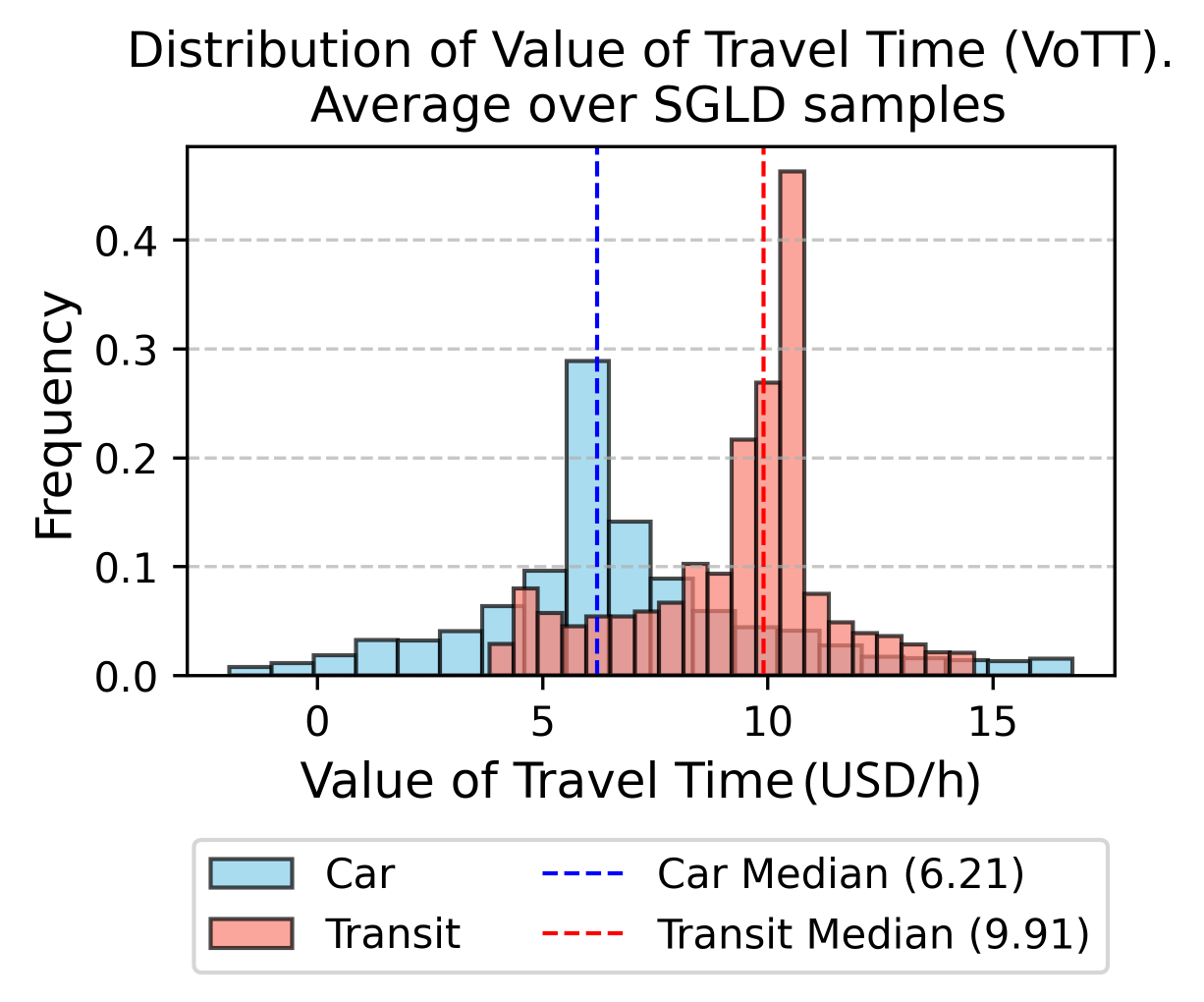}
        \caption{No $\ell_2$ penalty.}
        \label{fig:vott_skip_nn_mode_mode_no_l2}
    \end{subfigure}
     \hfill
    \begin{subfigure}[t]{0.3\textwidth}
        \centering
        \includegraphics[width=\linewidth]{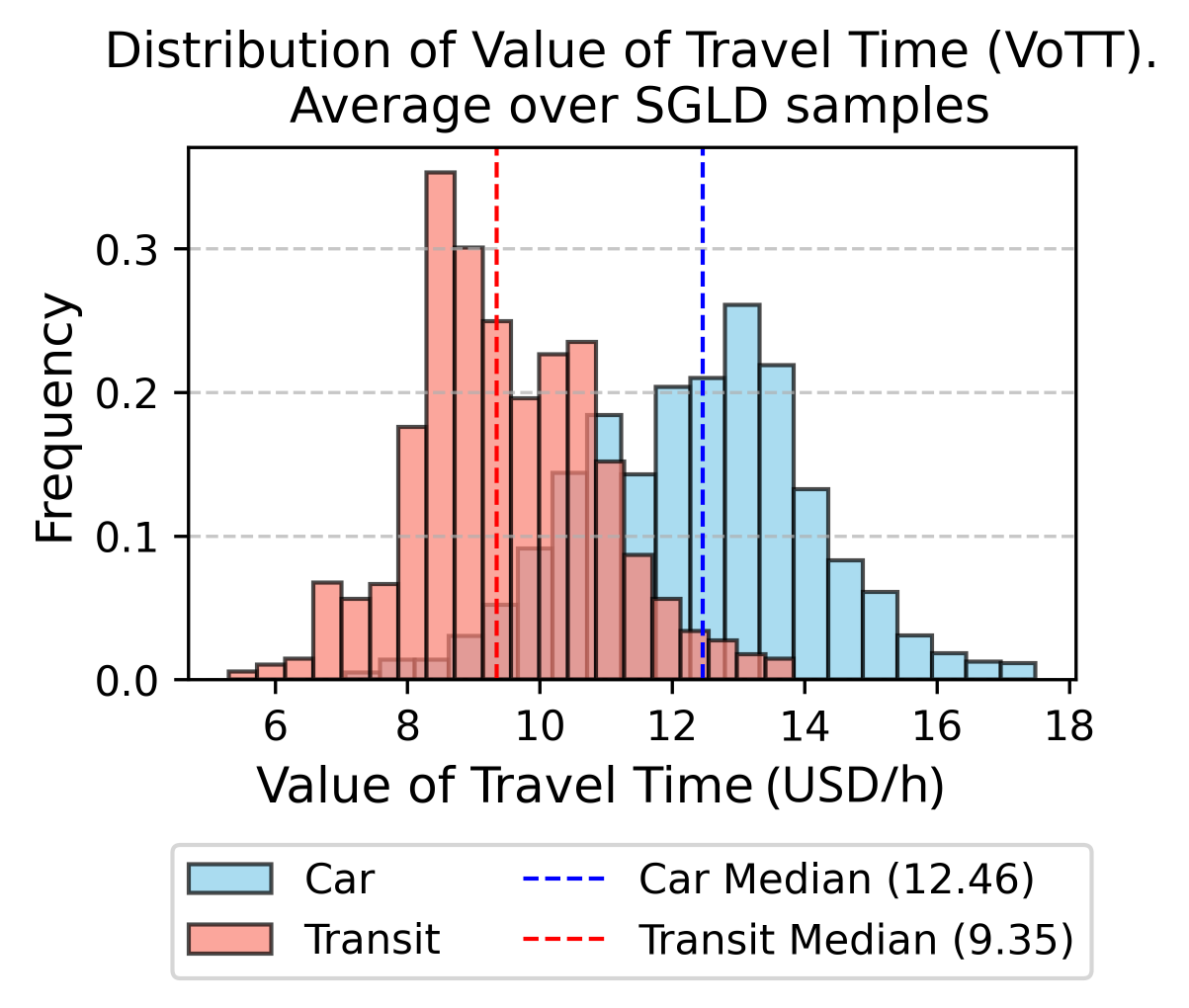}
        \caption{$\ell_2$ penalty from cross-validation applied to scale parameters.}
        \label{fig:vott_skip_nn_mode_mode_l2_cross_validation}
    \end{subfigure}
    \caption{Value of travel time savings (VOTT) for different $\ell_2$ penalties applied to the scale parameters in our model. Mode choice in New York City. Histograms are constructed by averaging over SGLD samples. The number of layers and hidden units for the embedding model and the non-linear components of the representative utilities is determined through cross-validation for each case.}
    \label{fig:vott_skip_nn}
\end{figure}

In Figure~\ref{fig:vott_skip_nn}, we show histograms across individuals in the sample for the average VOTT—computed across SGLD samples—under different $\ell_2$ regularization settings: several fixed a priori and one determined via cross-validation. For all configurations, the number of layers and hidden units in the embedding and nonlinear layers was selected via cross-validation. \\ 

We observe that, under all modeling conditions, our model estimates average VOTTs that align with behavioral intuition—a positive willingness to pay for a one-hour reduction in travel time—across the full sample, including the case with no $\ell_2$ regularization applied to the scale of the nonlinearities (Figure~\ref{fig:vott_skip_nn_mode_mode_no_l2}). For reference, in previous studies we have found VOTT estimates for car ranging from \$10 to \$35 per hour, and for transit from \$10 to \$14 per hour. Equally important, we show that the prior specification affects both the magnitude and the ranking of the VOTT estimates across alternatives, as seen by comparing the median VOTT values for car and transit across all configurations.\\

\begin{figure}[h!]
    \centering
    \begin{subfigure}[t]{0.45\textwidth}
        \centering
        \includegraphics[width=0.735\linewidth]{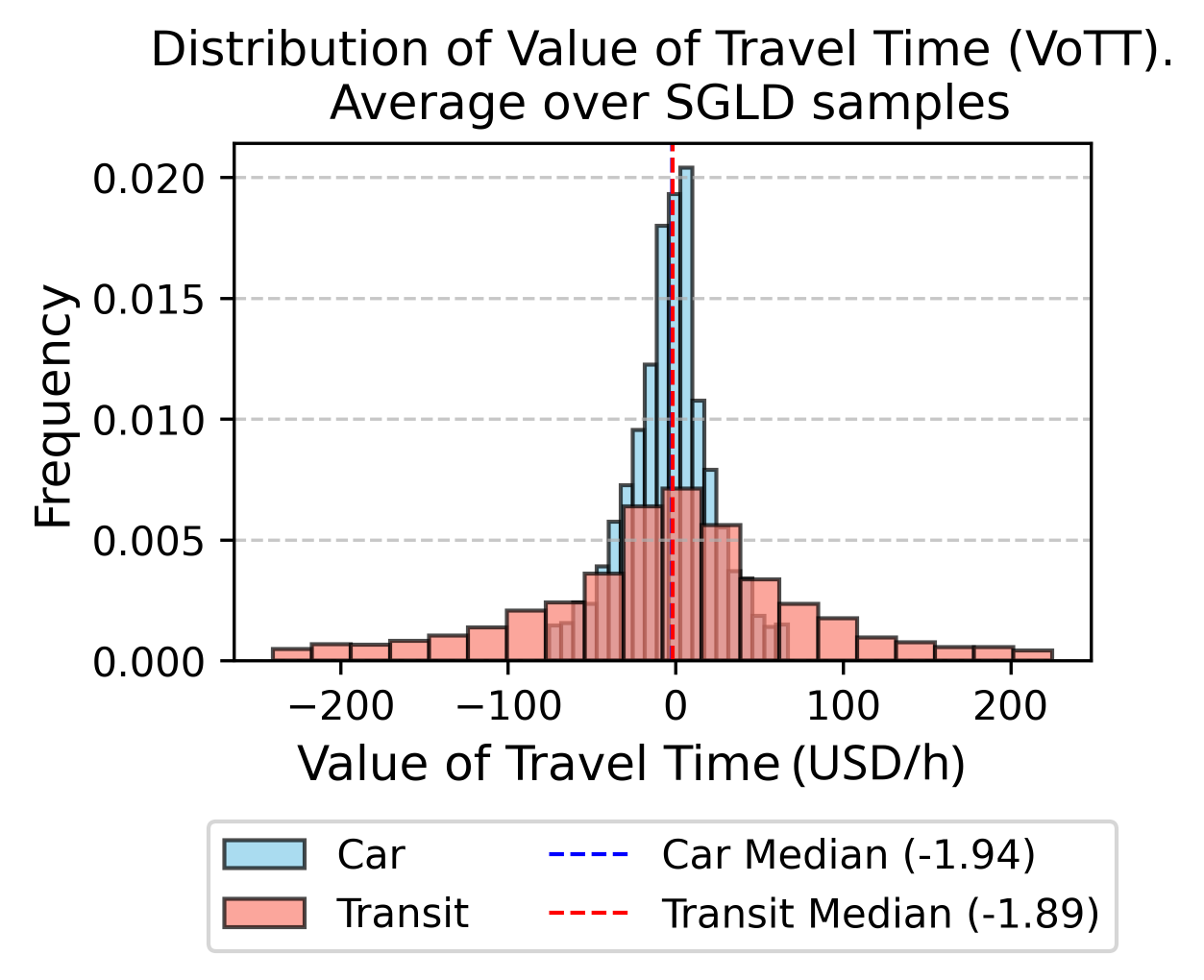}
        \caption{No weight decay (no $\ell_2$ penalty).}
        \label{fig:vott_fully_nn_mode_mode_no_l2}
    \end{subfigure}
    \hfill
    \begin{subfigure}[t]{0.45\textwidth}
        \centering
        \includegraphics[width=0.735\linewidth]{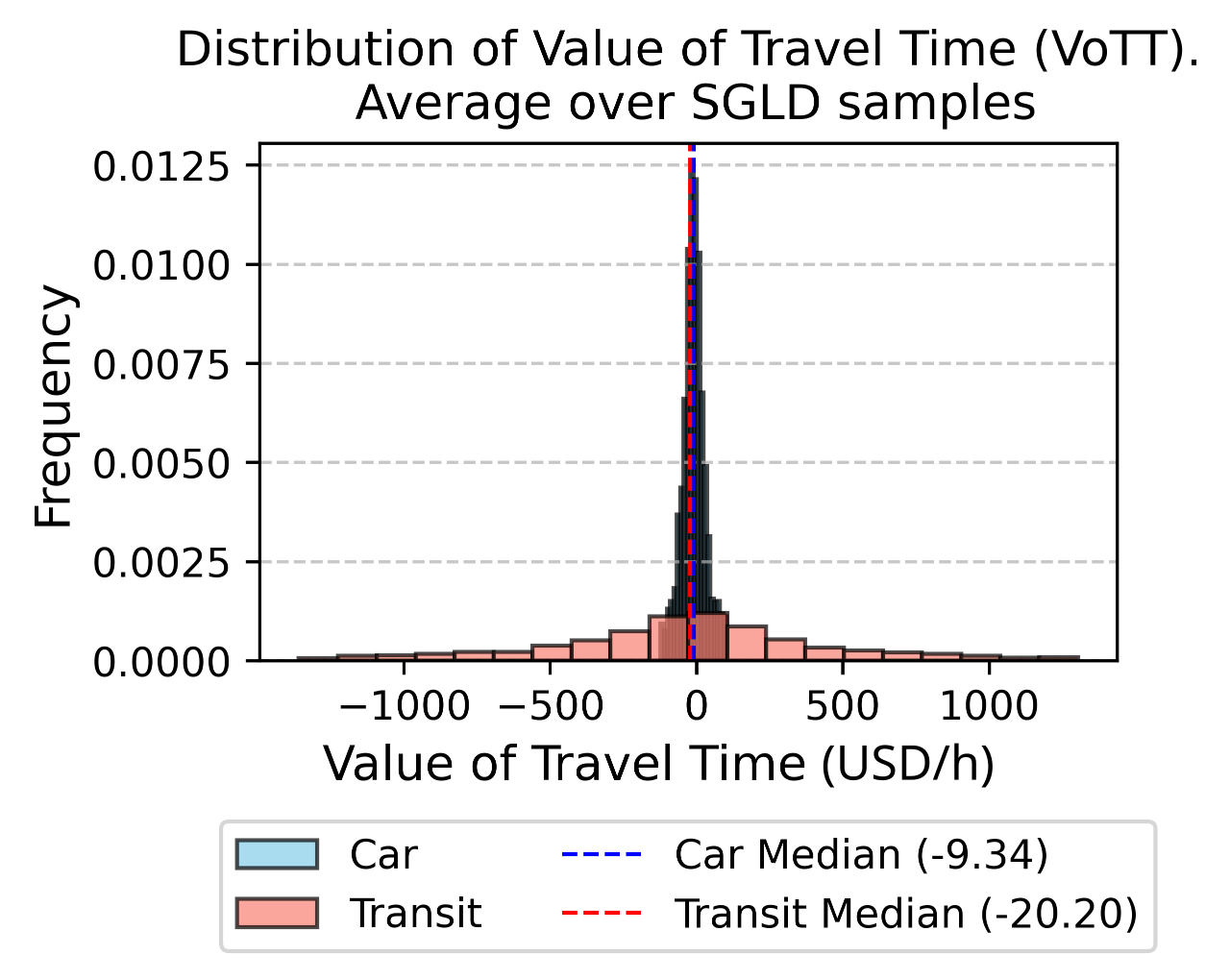}
        \caption{Weight decay from cross-validation ($\ell_2$ penalty applied to whole network).}
        \label{fig:vott_fully_nn_mode_mode_l2_cross_validation}
    \end{subfigure}
    \caption{Value of travel time savings (VOTT) for different $\ell_2$ penalties applied to the whole network. Mode choice in New York City. Histograms are constructed by averaging over SGLD samples. The number of layers and hidden units is determined from cross-validation for each case.}
    \label{fig:vott_fully_nn}
\end{figure}

On the other hand, we computed VOTT estimates using fully connected neural networks (with hidden layers and depth from cross-validation), both with and without $\ell_2$ penalties applied to the entire network. In both cases, we observe a significant proportion of negative willingness to pay for a one-hour reduction in travel time. The absolute values reach as high as \$200 for the model without regularization (Figure~\ref{fig:vott_fully_nn_mode_mode_no_l2}), and up to \$1000 for the model with regularization selected via cross-validation (Figure~\ref{fig:vott_fully_nn_mode_mode_l2_cross_validation}). For these models, even the median VOTT values across the population defy behavioral intuition by being negative. \\

In Table~\ref{tab:accuracy_mode}, we present the out-of-sample weighted accuracies for all models and $\ell_2$ penalization strategies. Our model achieves a performance level comparable to the state of the art while providing estimates that align with established behavioral knowledge. In contrast, although the fully connected neural network—implemented within a Bayesian framework using SGLD—attains similar predictive performance, the corresponding VOTT estimates are unreasonable.\\

\begin{table}[H]
    \centering
    \caption{Model out-of-sample accuracy mode choice in NYC. Predictions from the average over SGLD samples for latent utilities.}
    \begin{tabular}{cc|c}
    \hline
    Model & $\ell_2$ penalty & Accuracy (\%)\\
    \hline
    Our model & From cross validation & 78.15\\
    Our model & None & 75.62\\
    Our model & High for non-IIA scale parameter & 69.44\\
    Our model & High for non-linear scale parameters & 76.75\\
    Fully connected NN & From cross validation & 78.71 \\
    Fully connected NN & None & 76.18 \\
    Fully connected NN & High & 35.02 \\
    SOTA& - & 78.45 \tablefootnote{From the Skip-GNN presented by us in \cite{2025arXiv250309786V}.} \\
    \hline
    \end{tabular}
    \label{tab:accuracy_mode}
\end{table}

\subsection{Swiss Train Data}

For the stated preference dataset on Swiss train route choice, we computed the VOTT for the two route alternatives, as well as the out-of-sample prediction accuracy based on the average of the SGLD latent utilities. While previous studies have estimated VOTT values, few have reported out-of-sample predictive accuracies. However, we found a recent implementation by \cite{Li2024} that compares the predictive performance of ICLV models with that of discrete choice models enhanced with graph neural networks (GNNs) and data augmentation strategies with generative AI.\\

\begin{figure}[h!]
    \centering
    \begin{subfigure}[t]{0.3\textwidth}
        \centering
        \includegraphics[width=\linewidth]{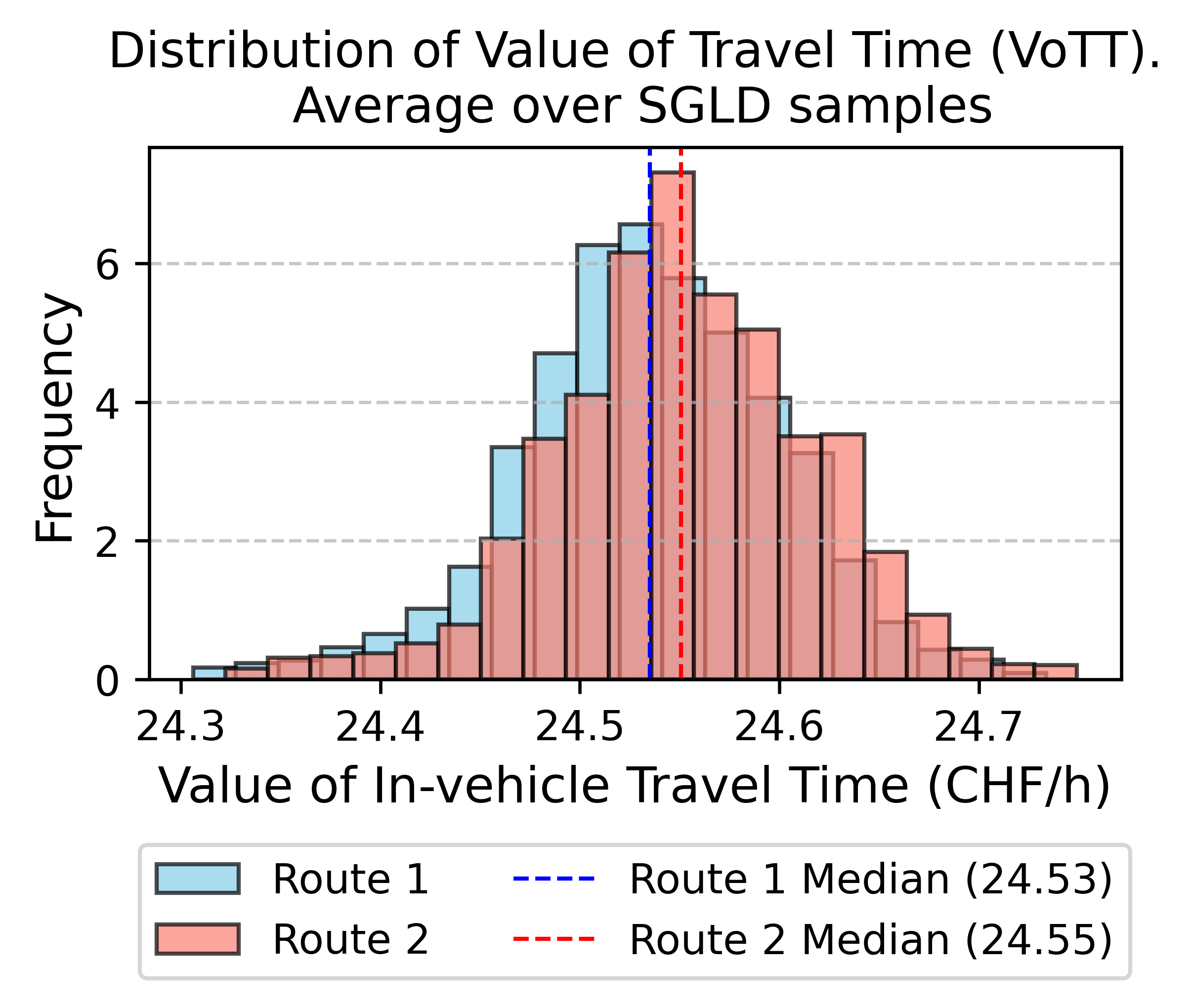}
        \caption{High $\ell_2$ penalty applied to non-linear scale parameters.}
        \label{fig:vott_skip_nn_swiss_mode_high_l2}
    \end{subfigure}
    \hfill
    \begin{subfigure}[t]{0.3\textwidth}
        \centering
        \includegraphics[width=\linewidth]{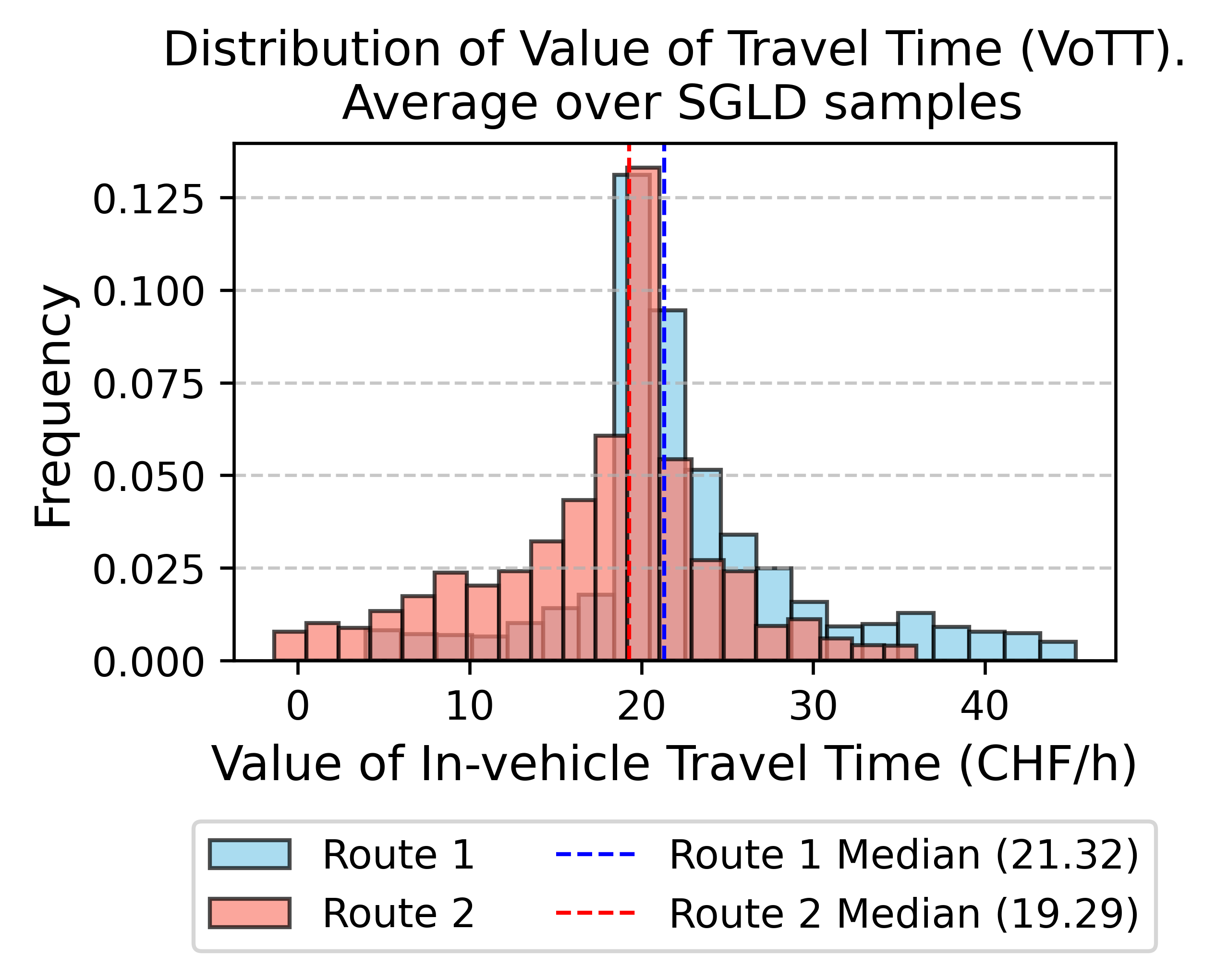}
        \caption{High $\ell_2$ penalty applied to non-linear non-IIA scale parameter. No penalty for non-linear IIA scale parameter.}
        \label{fig:vott_skip_nn_swiss_mode_high_l2_non_iia}
    \end{subfigure}
    \hfill
    \begin{subfigure}[t]{0.3\textwidth}
        \centering
        \includegraphics[width=\linewidth]{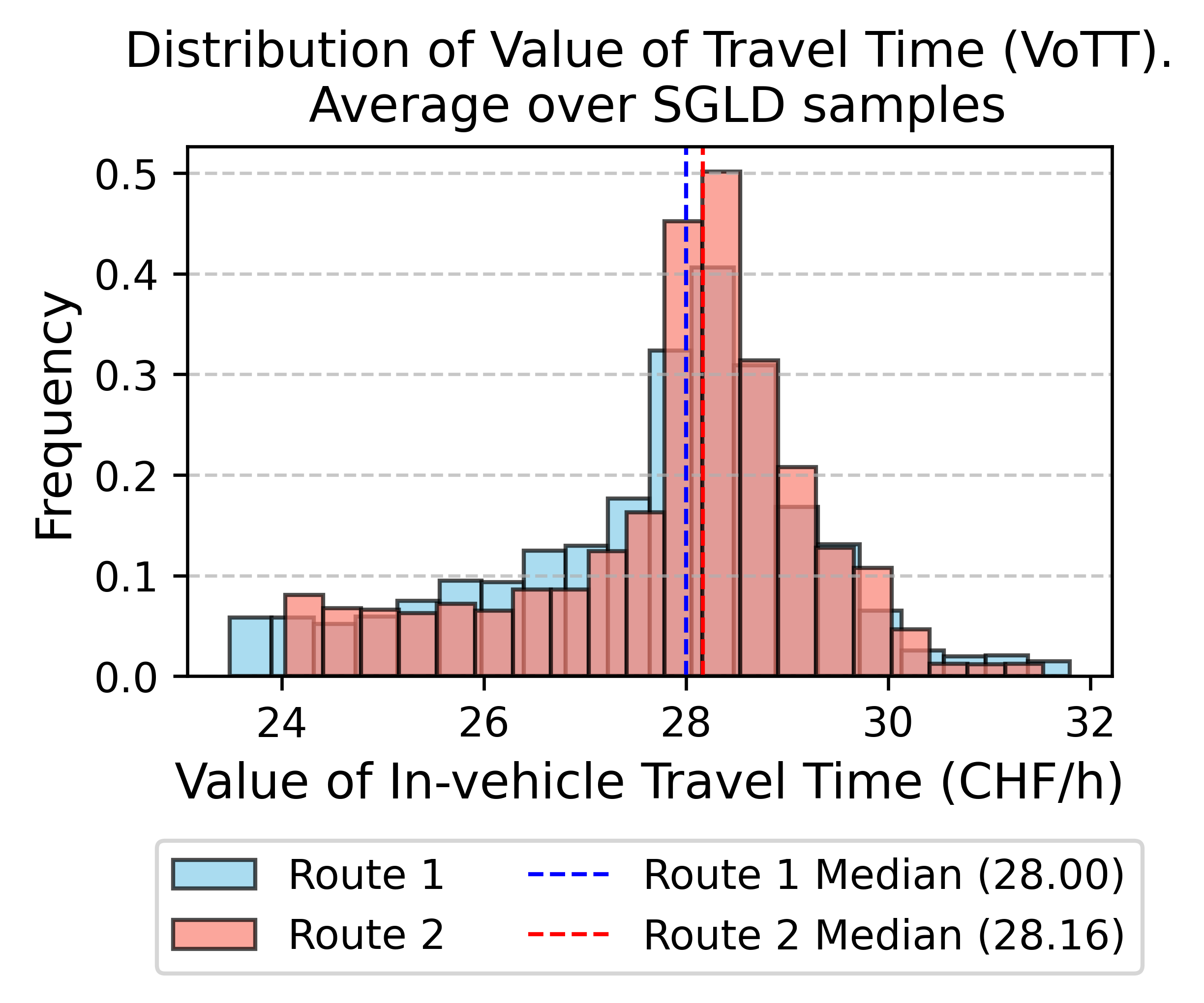}
        \caption{No $\ell_2$ penalty.}
        \label{fig:vott_skip_nn_swiss_mode_no_l2}
    \end{subfigure}
     \hfill
    \begin{subfigure}[t]{0.3\textwidth}
        \centering
        \includegraphics[width=\linewidth]{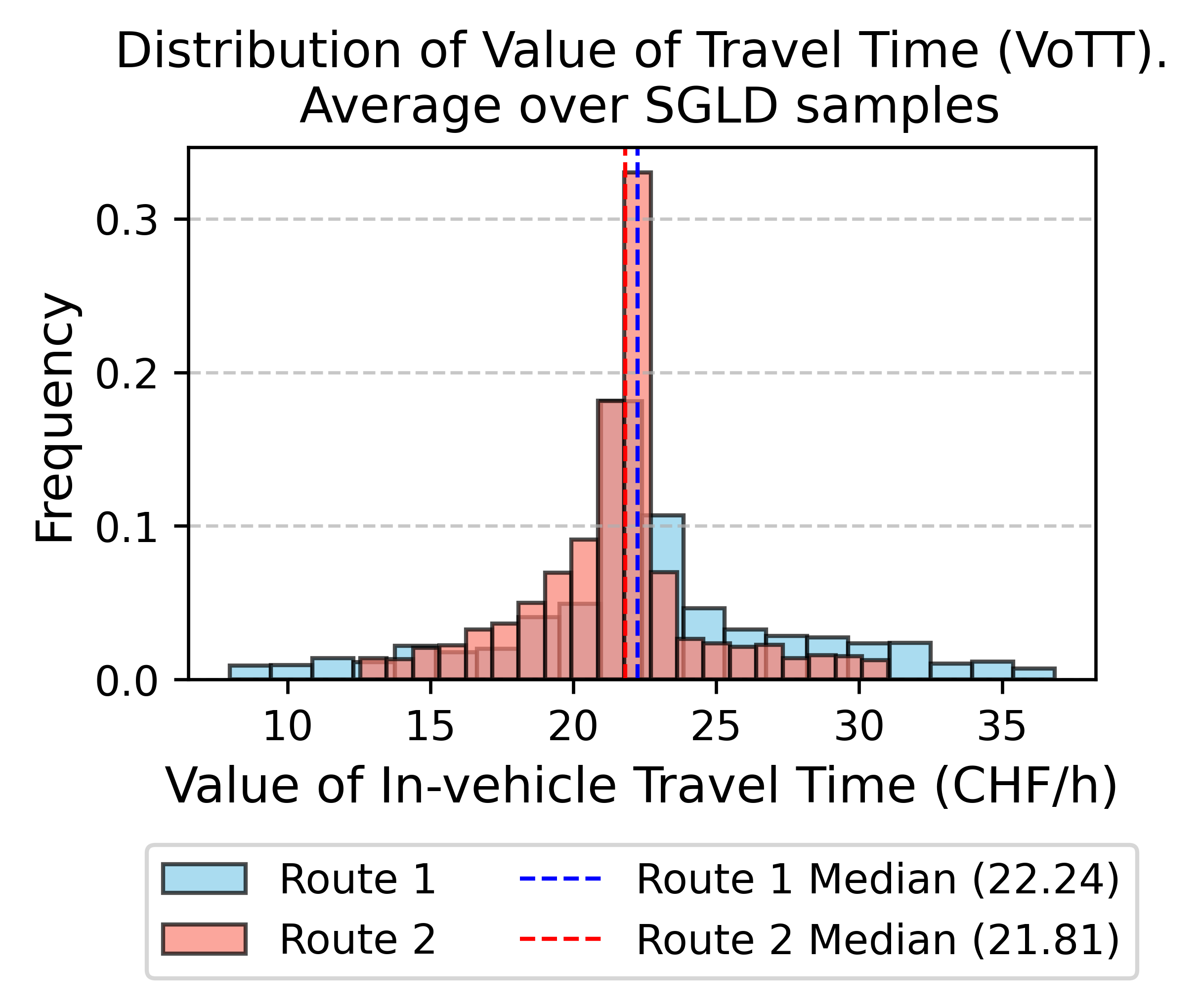}
        \caption{$\ell_2$ penalty from cross-validation applied to scale parameters.}
        \label{fig:vott_skip_nn_swiss_mode_l2_cross_validation}
    \end{subfigure}
    \caption{Value of in-vehicle travel time savings (VOTT) for different $\ell_2$ penalties applied to the scale parameters in our model. Swiss train route choice. Histograms are constructed by averaging over SGLD samples. The number of layers and hidden units for the embedding model and the non-linear components of the representative utilities is determined through cross-validation for each case.}
    \label{fig:vott_skip_nn_swiss}
\end{figure}

In Figure~\ref{fig:vott_skip_nn_swiss}, we present histograms of the average value of travel time savings (VOTT) across individuals in the sample, computed by averaging over SGLD samples, under various $\ell_2$ regularization settings. For all configurations, the number of layers and hidden units in both the embedding and nonlinear components was selected through cross-validation, following the same approach as in the NYC case study. Across all settings—including the case without $\ell_2$ regularization applied to the scale parameters of the nonlinearities (Figure~\ref{fig:vott_skip_nn_swiss_mode_no_l2})—our model produces average VOTT estimates that align with behavioral expectations, indicating a positive willingness to pay for a one-hour reduction in travel time. For reference, \citet{vrtic2002impact} report VOTT values ranging from 11.9 CHF/h for leisure trips to 52.4 CHF/h for business trips, with an overall average of 17.1 CHF/h for the full sample, regardless of trip purpose.\\

\begin{figure}[h!]
    \centering
    \begin{subfigure}[t]{0.45\textwidth}
        \centering
        \includegraphics[width=0.735\linewidth]{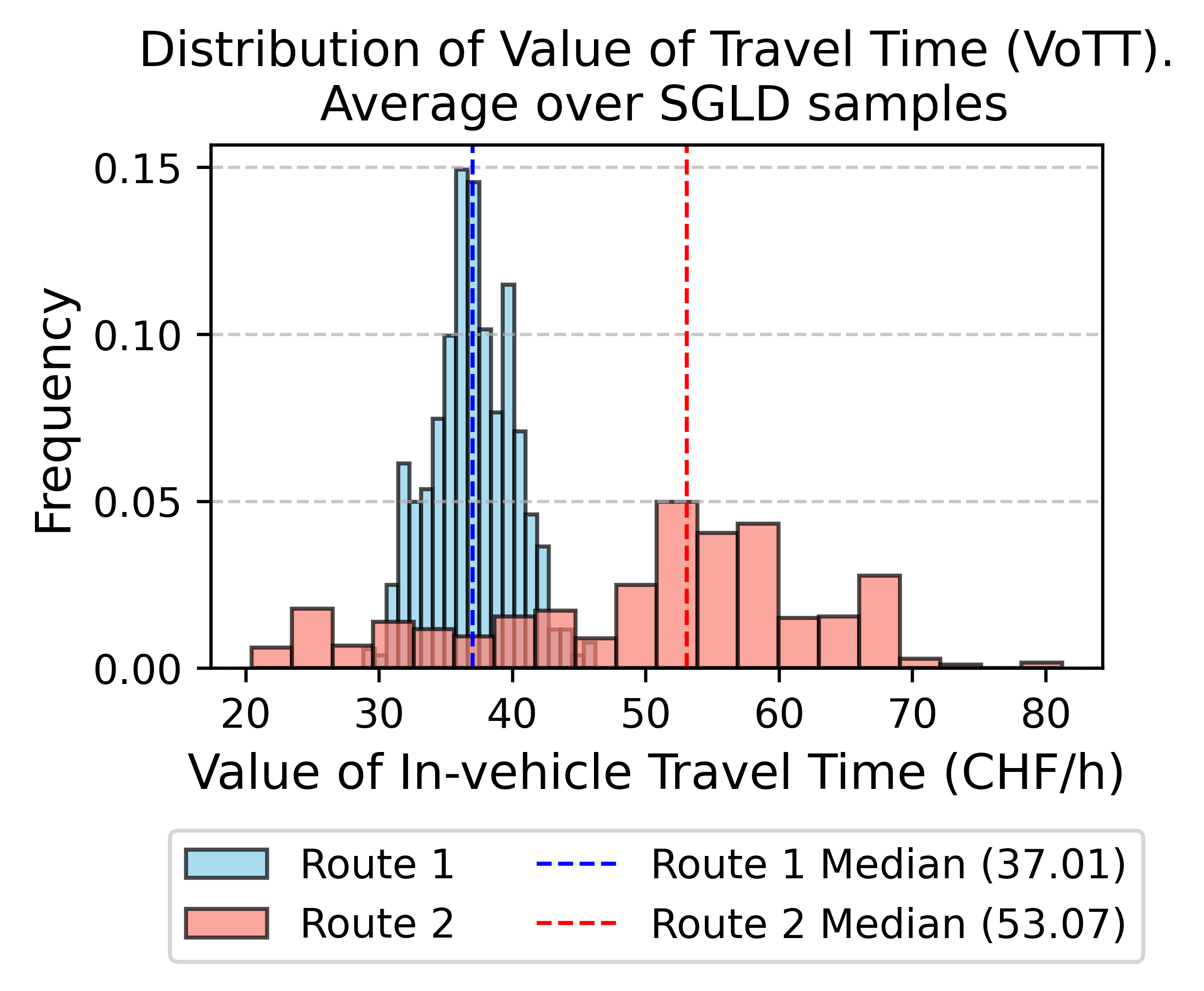}
        \caption{No weight decay (no $\ell_2$ penalty).}
        \label{fig:vott_fully_nn_swiss_mode_no_l2}
    \end{subfigure}
    \hfill
    \begin{subfigure}[t]{0.45\textwidth}
        \centering
        \includegraphics[width=0.735\linewidth]{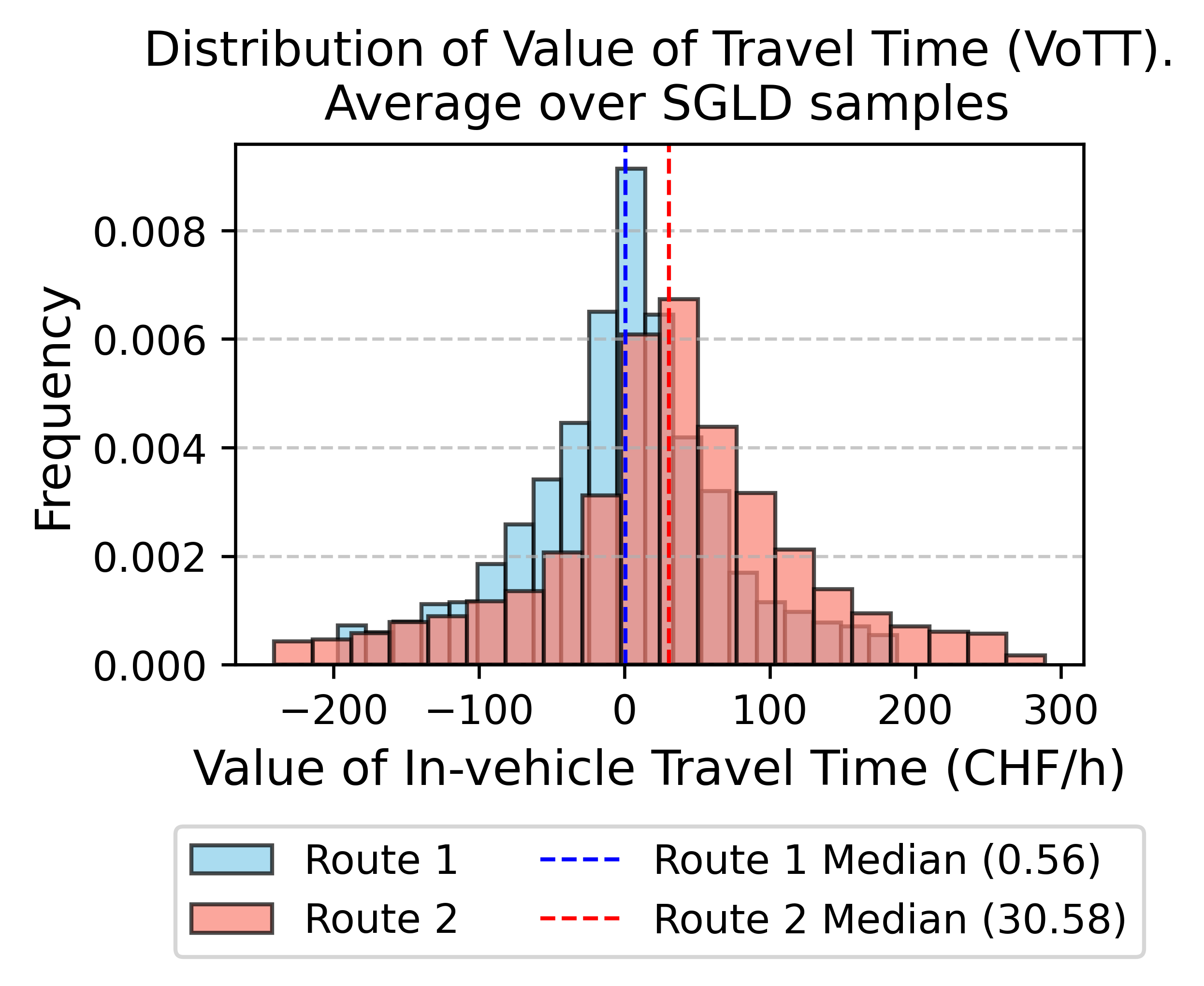}
        \caption{Weight decay from cross-validation ($\ell_2$ penalty applied to whole network).}
        \label{fig:vott_fully_nn_swiss_mode_l2_cross_validation}
    \end{subfigure}
    \caption{Value of travel time savings (VOTT) for different $\ell_2$ penalties applied to the whole network. Swiss train route choice. Histograms are constructed by averaging over SGLD samples. The number of layers and hidden units is determined from cross-validation for each case.}
    \label{fig:vott_fully_nn_swiss}
\end{figure}

Regarding the fully connected neural network, in Figure~\ref{fig:vott_fully_nn_swiss}, we show the VOTT histograms for the sample. As illustrated, the model without $\ell_2$ regularization produces positive estimates; however, the distributions appear to differ substantially in scale and location between the two routes, with values reaching up to 80~CHF/h. For the model with $\ell_2$ regularization determined via cross-validation, we observe similar results to those from the previous case study: a significant proportion of individuals have negative VOTT estimates, with absolute values exceeding 200~CHF/h.\\

Finally, in Table~\ref{tab:accuracy_mode_swiss}, we compare the out-of-sample accuracies of our model with state-of-the-art performance reported in the literature and that of a fully connected neural network. In this case study, we observe that our model—regardless of the penalization strategy—outperforms all other models considered, including the state-of-the-art benchmark (by six percentage points for two regularization strategies). Moreover, the fully connected neural network performs poorly under all regularization settings. \\

\begin{table}[H]
    \centering
    \caption{Model out-of-sample accuracy Swiss train route choice. Predictions from the average over SGLD samples for latent utilities.}
    \begin{tabular}{cc|c}
    \hline
    Model & $\ell_2$ penalty & Accuracy (\%)\\
    \hline
    Our model & From cross validation & 79.27\\
    Our model & None & 81.68 \\
    Our model & High for non-IIA scale parameter & 76.95 \\
    Our model & High for non-linear scale parameters & 81.58\\
    Fully connected NN & From cross validation & 53.34\\
    Fully connected NN & None & 60.43\\
    Fully connected NN & High & 50.00\\
    SOTA& - & 75.53\tablefootnote{From the GNN-MNL composite model presented in \cite{Li2024}. The reported metric in this paper is most likely raw accuracy. In our study, we rely on weighted accuracy, making the performance difference even more significant.} \\
    \hline
    \end{tabular}
    \label{tab:accuracy_mode_swiss}
\end{table}

\section{Conclusion}

In this paper, we have developed and implemented a general deep learning architecture that combines the expressive power of neural networks with the ability to collapse to behaviorally informed hypotheses when the data does not support more complex representations. Our model consists of a behaviorally informed component, an embedding network over shared inputs (such as socio-demographics), a nonlinear IIA component, and a nonlinear non-IIA component. We proposed a straightforward two-step learning procedure that guides the SGLD sampler toward regions of the parameter space where the magnitudes of the behaviorally informed coefficients are high. This approach ensures that the nonlinear components gain importance only when they offer significant improvements in the posterior relative to the informed part of the model. \\

To our knowledge, this paper is the first to explore Bayesian deep learning for discrete choice modeling—beyond the brief introduction we previously presented in \cite{2025arXiv250309786V}. By leveraging SGLD, we are able to provide credible intervals for economic quantities such as marginal rates of substitution (e.g., values of travel time), and to make predictions using distributions over parameter sets rather than point estimates, which are prone to overfitting when the model is underspecified. With our model design and two-step training procedure—which combines convergence to behaviorally sound hypotheses with posterior sampling over more complex representations—we effectively address key concerns raised by the discrete choice community regarding the application of deep learning: model architectures that lack behavioral interpretability, unstable parameter estimates, and the absence demonstrations on how to quantify and represent uncertainty.\\

We tested our model empirically using a simulation study, a revealed preference study on mode choice in New York City, and a stated preference study on train route choice in Switzerland. In the simulation study, we computed empirical coverage for marginal rates of substitution as well as posterior prediction bands for the latent utilities. Our model attained empirical coverages that matched the credibility level, whereas the fully connected neural network underperformed in low-data settings, and the model based solely on the behaviorally informed hypothesis failed to achieve adequate empirical coverage at any dataset size. Regarding the latent utilities, we observed consistent patterns: our model closely approximated the nonlinear utilities and produced posterior bands that narrowed with increasing dataset size—indicating that our SGLD implementation within the two-step training procedure reliably captured epistemic uncertainty. Additionally, our model achieved predictive performance equivalent to—or better than—that of the other two models across all dataset sizes, demonstrating that it can deliver competitive predictive accuracy while also providing reliable point and interval estimates.\\ 

In the revealed and stated preference case studies, we estimated the value of travel time savings for each individual in the sample using posterior samples obtained from SGLD. We found that our model produced VOTT estimates that aligned with behavioral intuition, regardless of the prior imposed through the $\ell_2$ regularization constants on the scale of the nonlinear components. In contrast, the fully connected neural network produced a significant proportion of negative VOTT estimates and yielded extremely large absolute values. Additionally, our model achieved competitive out-of-sample predictive accuracy, outperforming the state of the art by six percentage points in the Swiss route choice dataset.\\

In this study, we imposed priors only on the nonlinear scales of our model through $\ell_2$ regularization constants. However, the modeling strategy is flexible and allows for alternative prior distributions on the scales that are not limited to Gaussian assumptions. Future research should further investigate the impact of prior specification on both model performance and the computational complexity of the two-step training procedure. Additionally, although we implemented SGLD for the sampling stage of our training process, other inference techniques from the Bayesian deep learning literature are worth exploring. We are confident that this paper will encourage greater attention to the role of deep learning models—and potentially other flexible, differentiable models—in discrete choice analysis, not only for improving predictive performance but also for enabling more reliable inference.

\section*{Acknowledgments}
 This research was supported by the National Science Foundation Award No. SES-2342215. We are also thankful for the financial support provided by the Fulbright Scholarship Program, which is sponsored by the U.S. Department of State, the Colombian Fulbright Commission, and the Colombian Science Ministry. This project is solely the responsibility of the authors and does not necessarily represent the official views of the Fulbright Program, the U.S. government, or the Colombian government.

\newpage
\bibliographystyle{elsarticle-num-names} 
\bibliography{biblio}

\end{document}